\documentclass[10pt,twocolumn,letterpaper]{article}
\usepackage{authblk}
\usepackage[pagenumbers]{cvpr} 
\usepackage[title]{appendix}
\usepackage{graphicx}
\usepackage{amsmath}
\usepackage{amssymb}
\usepackage{booktabs}

\usepackage{algorithmic}
\usepackage{algorithm}
\usepackage{bm}
\usepackage{multirow}
\usepackage{color}
\usepackage{bbm}
\usepackage{caption}
\usepackage[table]{xcolor}

\usepackage[pagebackref,breaklinks,colorlinks]{hyperref}

\usepackage[capitalize]{cleveref}
\crefname{section}{Sec.}{Secs.}
\Crefname{section}{Section}{Sections}
\Crefname{table}{Table}{Tables}
\crefname{table}{Tab.}{Tabs.}
\usepackage{caption}

\author[1]{Pengchong Qiao}
\author[1]{Zhidan Wei}
\author[1]{Yu Wang}
\author[2]{Zhennan Wang}
\author[2]{Guoli Song}
\author[2]{Fan Xu}
\author[3]{Xiangyang Ji}
\author[3]{Chang Liu}
\author[1,2]{Jie Chen}
{\affil[1]{Peking University}\affil[2]{Pengcheng Laboratory}\affil[3]{Tsinghua University}}

\begin{document}

\title{Fuzzy Positive Learning for Semi-supervised Semantic Segmentation}

%
\maketitle

\begin{abstract}
Semi-supervised learning (SSL) essentially pursues class boundary exploration with less dependence on human annotations.
Although typical attempts focus on ameliorating the inevitable error-prone pseudo-labeling, we think differently and resort to exhausting informative semantics from multiple probably correct candidate labels.
In this paper, we introduce Fuzzy Positive Learning (FPL) for accurate SSL semantic segmentation in a plug-and-play fashion, targeting adaptively encouraging fuzzy positive predictions and suppressing highly-probable negatives.
Being conceptually simple yet practically effective, FPL can remarkably alleviate interference from wrong pseudo labels and progressively achieve clear pixel-level semantic discrimination.
Concretely, our FPL approach consists of two main components, including fuzzy positive assignment (FPA) to provide an adaptive number of labels for each pixel and fuzzy positive regularization (FPR) to restrict the predictions of fuzzy positive categories to be larger than the rest under different perturbations.
Theoretical analysis and extensive experiments on Cityscapes and VOC 2012 with consistent performance gain justify the superiority of our approach.
%

\end{abstract}

\section{Introduction}
\label{sec:intro}
Semantic segmentation models enable accurate scene understanding~\cite{xiao2018unified, behley2019semantickitti, minaee2021image} with the help of fine pixel-level annotations.
Yet, collecting labeled segmentation datasets is time-consuming and labor-costing~\cite{cordts2016cityscapes}.
Considering unlabeled data are annotation-free and easily accessible, semi-supervised learning (SSL) is introduced into semantic segmentation~\cite{zou2020pseudoseg,yuan2021simple,ouali2020semi,chen2021semi,zhong2021pixel,wang2022semi} to encourage the model to generalize better on unseen data with less dependence on artificial annotations.


\begin{figure}[htbp]
\centerline{\includegraphics[scale=0.32]{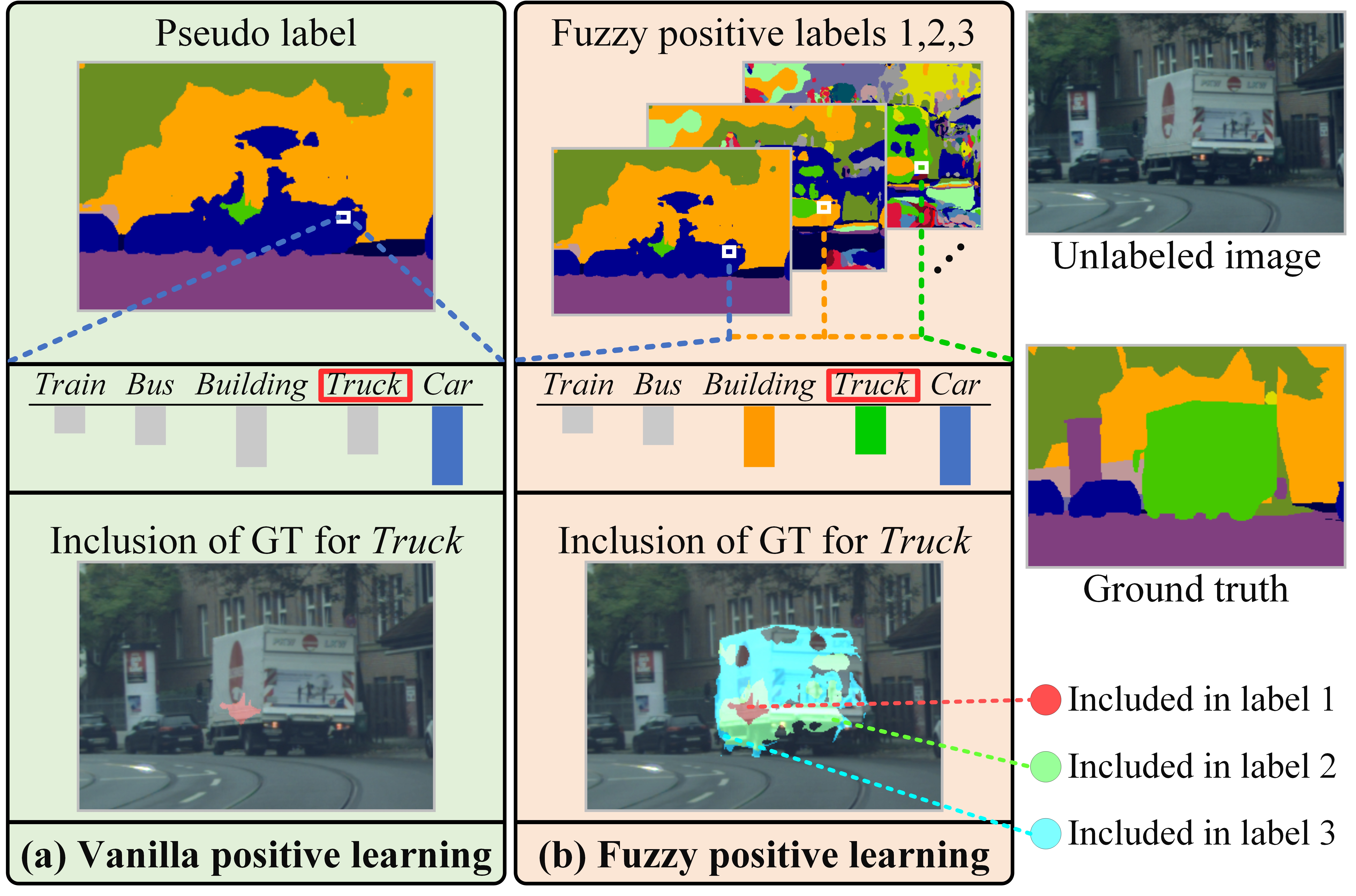}}
\caption{
(a) Existing methods using pseudo label to utilize unlabeled data.
(b) The proposed FPL that provides multiple fuzzy positive labels for each pixel to utilize unlabeled data.
The example of `Truck' shows that our method covers ground truth (GT) more comprehensively than vanilla positive learning.
}
\label{fig:intro}
\vspace{-1.5em}
\end{figure}

The semi-supervised segmentation task faces a scenario where only a subset of training images are assigned segmentation labels while the others remain unlabeled.
Current state-of-the-art (SOTA) methods utilize unlabeled data via consistency regularization, which aims to obtain invariant predictions for unlabeled pixels under various perturbations~\cite{zou2020pseudoseg,yuan2021simple,ouali2020semi,chen2021semi}.
Their general paradigm is to use the pseudo label generated under weak (or none) perturbations as the learning target of predictions under strong perturbations.
%
Though achieving promising results, errors are inevitable in the pseudo label used in these methods, misguiding the training of their models~\cite{li2017learning,oliver2018realistic}.
An intuitive example is that some pixels may be confused in categories with similar semantics.
As Fig.~\ref{fig:intro} (a), some pixels belonging to `Truck' are wrongly classified into the `Car' category (e.g., white boxed pixel).
To mitigate this problem, typical methods focus on ameliorating the learning of pseudo labels by filtering low-confidence pseudo labels out~\cite{sohn2020fixmatch,zou2020pseudoseg, hu2021semi,lai2021semi,zhong2021pixel} and generating pseudo labels more accurately~\cite{liu2022perturbed,yang2022st++,fan2022ucc,kwon2022semi}.
However, the semantics of ground truth buried in other unselected labels are ignored in existing methods.
In this paper, we propose \textbf{Fuzzy Positive Learning} (FPL), a new SSL segmentation method that exhausts informative semantics from
multiple probably correct candidate labels.
We name these labels ``fuzzy positive'' labels since each of them has the probability to be the ground truth.
As shown in Fig.~\ref{fig:intro} (b), our fuzzy positive labels cover the ground truth more comprehensively, facilitating our FPL to exploit the semantics of ground truth better.
Extending learning from one pseudo label to learning from multiple fuzzy positive labels is not a simple implementation, which contains two pending issues.
One is how to provide an adaptive number of labels for each pixel.
And the other one is how to exploit the possible GT semantics from fuzzy positive labels.
For these two issues, a fuzzy positive assignment (FPA) algorithm is first proposed to select which labels should be appended to the fuzzy positive label set of each pixel.
Afterward, a fuzzy positive regularization (FPR) is developed to regularize the predictions of fuzzy positive categories to be larger than the predictions of the rest negative categories under different perturbations.

Our FPL achieves consistent performance gain on Cityscapes and Pascal VOC 2012 datasets using CPS~\cite{chen2021semi} and AEL~\cite{hu2021semi} as baselines.
Moreover, we theoretically and empirically analyze that the superiority of FPL lies in revising the gradient of learning ground truth when pseudo-labels are wrongly-assigned.
Our main contributions are:
\begin{itemize}
  \item FPL provides a new perspective for SSL segmentation, that is, learning informative semantics from multiple fuzzy positive labels instead of only one pseudo label.
  \item A fuzzy positive assignment is proposed to provide an adaptive number of labels for each pixel. Besides, a fuzzy positive regularization is developed to learn the semantics of ground truth from fuzzy positive labels.
  \item FPL is easy to implement and could bring stable performance gains on existing SSL segmentation methods in a plug-and-play fashion.
\end{itemize}

\section{Related Work}
\subsection{Semi-supervised Learning}
Modern SSL classification approaches typically learn semantics from unlabeled data by introducing techniques of entropy minimization and consistency regularization.
Entropy minimization enforces the predicted probability distribution to be sharp by training upon pseudo labels~\cite{lee2013pseudo, mcelreath2018statistical, xie2019unsupervised, berthelot2019mixmatch, berthelot2019remixmatch,sohn2020fixmatch}.
On the other hand, consistency regularization aims to obtain prediction invariance under various perturbations, including input perturbation~\cite{miyato2018virtual, xie2019unsupervised, sohn2020fixmatch}, feature perturbation~\cite{ouali2020semi}, network perturbation~\cite{tarvainen2017mean, pham2021meta, ke2019dual,feng2022dmt}, \textit{etc}.
Variants of their combination have achieved great success~\cite{sohn2020fixmatch, zhang2021flexmatch, xu2021dash, pham2021meta, wei2021crest}, whose core inspiration is computing consistency regularization via pseudo labeling.

\subsection{Semi-supervised Semantic Segmentation}
Semi-supervised semantic segmentation methods benefit from the development of general semi-supervised learning, which could be also roughly divided into two types of approaches: consistency regularization based methods~\cite{kim2020structured,french2019semi,ke2020guided,ouali2020semi} and entropy-minimization based methods~\cite{chen2020naive,zhu2021improving,ibrahim2020semi,feng2020semi,mittal2019semi,mendel2020semi}.
More recently, SOTA semi-supervised segmentation methods combine both two technologies together to train their models.
PseudoSeg~\cite{zou2020pseudoseg}, AEL~\cite{hu2021semi}, UCC~\cite{fan2022ucc} and Jianglong Yuan et al.~\cite{yuan2021simple} propose to use the pseudo label generated from weak augmented image to supervise the prediction of strong augmented image.
CPS~\cite{chen2021semi} designs a mutual learning mechanism that trains two student models with pseudo labels from each other.
PC$^{2}$Seg~\cite{zhong2021pixel} proposes a negative sampling technique to provide reliable negative samples for SSL segmentation.
Different from existing methods, we propose for the first time to exploit the informative semantics of unlabeled data from multiple fuzzy positive labels, resulting in less interference from wrong pseudo labels and accurate segmentation.


\textbf{Pseudo-label learning} is the key technology in current SSL segmentation methods, but it has a limitation in that wrong pseudo labels mislead the training of SSL models.
Typical approaches design filter-out mechanisms to use only high-confidence pseudo-labels for training~\cite{hu2021semi,fan2022ucc,zou2020pseudoseg,lai2021semi,zhong2021pixel} and develop complex training mechanisms to predict accurate pseudo-labels~\cite{liu2022perturbed,yang2022st++,fan2022ucc,kwon2022semi}.
Apart from the above methods, U$^{2}$PL~\cite{wang2022semi} introduces the idea of negative learning into SSL segmentation, which has similarities to our FPL.
It thinks uncertain pixels usually get confused among only a few classes.
Hence, it uses uncertain pixels as negative samples for those unlikely classes.
We analyze that our FPL and negative learning have mathematically different optimization objectives.
That is, negative learning implicitly maximizes only the prediction of the pseudo-label, while our FPL learns all fuzzy positive labels.
(cf. Appendix).

\section{Method}\label{sec:method}

\subsection{Preliminaries}\label{sec: Preliminaries}
\textbf{Overview:}
For the SSL segmentation task, we have a small labeled dataset $D_{l} = \{(x_{l},y_{l})\}_{l=1}^{L}$ and a large unlabeled dataset $D_{u} = \{x_{u}\}_{u=1}^{U}$, where $L$ is the size of the labeled dataset, and $U$ is the size of the unlabeled dataset ($L\ll U$).
The $x_{l},y_{l},x_{u}$ are the image and label of the $l$-th labeled data and the image of the $u$-th unlabeled data, respectively.
The purpose of SSL segmentation is to learn the parameters $\theta$ of a segmentation model $\mathbb{F}(\bullet;\theta)$ by optimizing a loss function that contains both supervised and unsupervised loss:
\begin{equation}
  \begin{aligned}
    \mathcal{L} = \frac{1}{L}\sum_{l=1}^{L}\mathcal{L}^{sup}(\mathbb{F}(x_{l};\theta))+\frac{\beta}{U}\sum_{u=1}^{U} \mathcal{L}^{uns}(\mathbb{F}(x_{u};\theta)),
  \end{aligned} \label{eq:original_object}
\end{equation}
where $\mathcal{L}^{sup}$ and $\mathcal{L}^{uns}$ are supervised loss and unsupervised loss, and $\beta$ is a regularization weight.

\begin{figure*}[htbp]
\centerline{\includegraphics[scale=0.315]{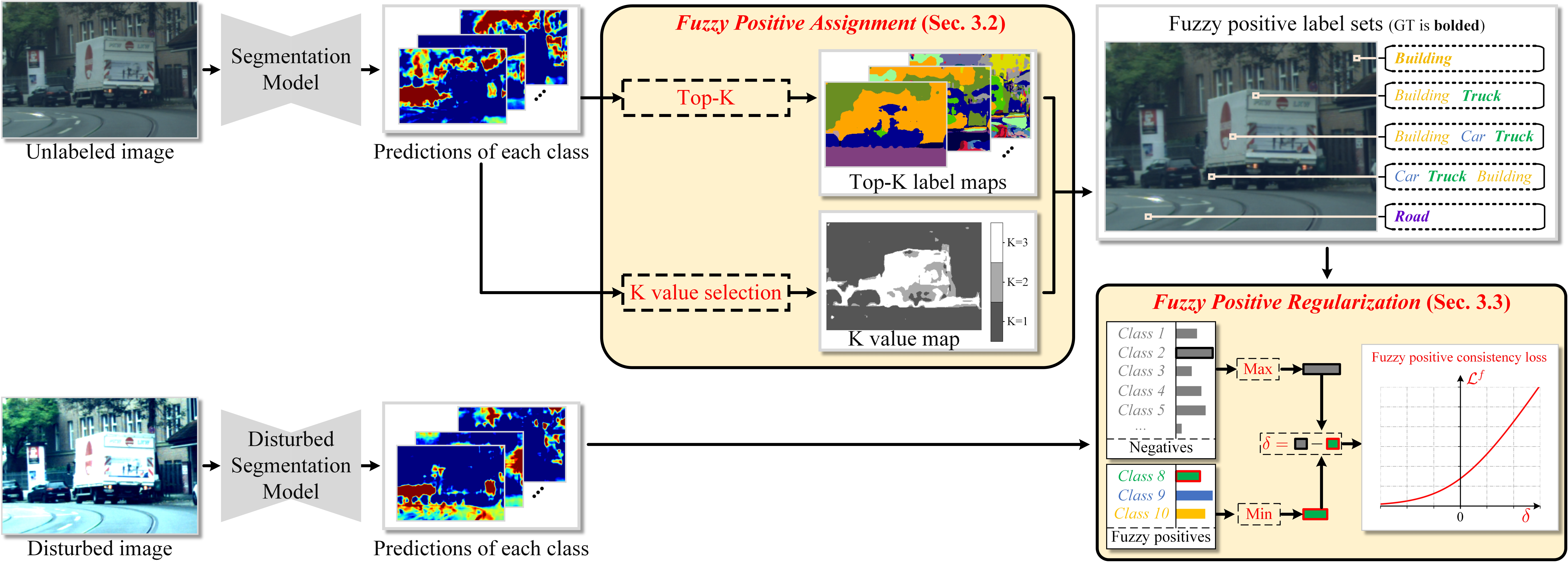}}
\caption{
\textbf{Pipeline illustration of our FPL}, where FPA densely allocates multiple labels as a fuzzy positive label set for each pixel, while FPR enforces the discrimination of the fuzzy positive assigns with the rest negative labels to facilitate more reliable semantic generalization.
}
\label{fig:framework}
\vspace{-1.0em}
\end{figure*}

In current SOTA methods~\cite{zou2020pseudoseg,chen2021semi,yuan2021simple,hu2021semi,kwon2022semi,fan2022ucc,liu2022perturbed}, the unsupervised loss in Eq.~\ref{eq:original_object} is formulated as the cross-entropy loss between model predictions and pseudo labels, which are also predicted by their models. The paradigm is:
\begin{equation}
\begin{aligned}
\overline{\bm{y}_{u}} &= \mathbbm{1}(\arg \max(\mathbb{F}(x_{u}))),~~~\bm{z}_{u} = \hat{\mathbb{F}}(\hat{x}_{u})\\
\mathcal{L}_{u}^{v} &= \frac{1}{S}\sum_{s=1}^{S} \mathcal{L}^{v}_{us} (\bm{z}_{us},\overline{\bm{y}_{us}}) \\
&= \frac{1}{S}\sum_{s=1}^{S} \sum_{c=1}^{C} -\overline{y_{us}^{c}}\log (\frac{\exp(z_{us}^{c})}{\sum_{n=1}^{C} \exp(z_{us}^{n})}),
\end{aligned} \label{eq:top1 consistency loss}
\end{equation}
where the $\overline{\bm{y}_{u}}$ is the one-hot encoding of the pseudo label generated from a segmentation model $\mathbb{F}$, and $\mathbbm{1}$ is the one-hot-encoding function.
The $\bm{z}_{u}$ is the prediction vector from disturbed model $\hat{\mathbb{F}}$ with disturbed input $\hat{x}_{u}$.
The disturbed model is often realized by adding dropout layers~\cite{laine2016temporal, ouali2020semi} into the model structure, or injecting random noises into the feature maps~\cite{ouali2020semi, liu2022perturbed}.
And the disturbed input is usually realized by data augmentations~\cite{zou2020pseudoseg,chen2021semi,yuan2021simple,hu2021semi}.
The $S$ is the number of pixels in image $x_{u}$ and $C$ is the number of categories, and $\overline{y_{us}^{c}}$ and $z_{us}^{c}$ are the elements of $\overline{\bm{y}_{u}}$ and $\bm{z}_{u}$ for the $c$-th class of the $s$-th pixel.
This vanilla positive loss $\mathcal{L}^{v}_{u}$ has only one learning target, the pseudo label.



\textbf{Motivation:}
By the definition of $\mathcal{L}^{v}_{us}$, its gradient with respect to the prediction $\bm{z}_{us}$ in backpropagation is computed as:
\begin{equation}
\begin{aligned}
&\frac{\partial \mathcal{L}^{v}_{us}}{\partial z_{us}^{c}}=
\left\{\begin{aligned} &p_{us}^{c}-1,~~~if~~\overline{y_{us}^{c}} = 1, \\
&p_{us}^{c},~~~~~~~~~else, \end{aligned}\right.
\end{aligned} \label{eq:gradient of top1 consistency loss}
\end{equation}
where the $p_{us}^{c} = \frac{\exp(z_{us}^{c})}{\sum_{n=1}^{C} \exp(z_{us}^{n})}$ is the predicted probability for the $c$-th class computed by softmax.
According to the gradient descent algorithm~\cite{rumelhart1985learning}, the prediction $z_{us}^{c}$ for category $c$ will increase if its gradient is less than 0, and vice versa.
In other words, only the prediction for the pseudo label category  ($\overline{y_{us}^{c}}=1$) is optimized to increase, and the predictions for other categories ($\overline{y_{us}^{c}}=0$) are optimized to decrease.
This shows that when the pseudo label is correct, vanilla positive learning encourages the prediction for the ground truth and suppresses the predictions for other categories, thus effectively utilizing unlabeled data.
However, once the pseudo-label is assigned incorrectly, the training of the SSL model will be misled since the prediction of ground truth is suppressed.

To reduce interference from wrong pseudo labels, we propose an FPL to exploit informative semantics from unlabeled data via multiple fuzzy positive labels, as shown in Fig.~\ref{fig:framework}.
Concretely, in Sec.~\ref{sec: candidate label proposal}, we propose a fuzzy positive assignment (FPA) algorithm, which assigns the top-K predicted categories of each pixel as its fuzzy positive labels, where K is computed according to our elaborate K value selection strategy.
In Sec.~\ref{sec: pessimistic consistency loss}, we develop a fuzzy positive regularization (FPR), which enables our model to exploit the possible ground truth in the fuzzy positive label set by regularizing the predictions of fuzzy positive categories to be larger than the rest negative categories.



\subsection{Fuzzy Positive Assignment}\label{sec: candidate label proposal}
The assignment of fuzzy positive labels determines from which our FPL exploits the semantics of ground truth.
To provide an adaptive number of labels for each pixel,
we first propose to choose the categories with top-K predicted probabilities as fuzzy positive labels since high-confidence predictions are prone to be correct~\cite{berthelot2019mixmatch}.
We then design an easy but effective K value selection strategy to adaptively determine the K value for each pixel, as shown in Alg.~\ref{alg:K value selection strategy}.
Specifically, we set a hyperparameter $T$ that represents
the upper bound of cumulative probability.
For each pixel, we compute the cumulative probability of its
top-n predicted categories and record the value of $n$ where the cumulative probability exceeds $T$ for the first time.
Finally, the K value for this pixel is set as $\max(n-1, 1)$.

\begin{algorithm}[tb]
   \caption{K value selection strategy}
   \label{alg:K value selection strategy}
\begin{algorithmic}
    \STATE \textbf{Input:} sorted prediction $\textbf{p}=(p^{1},p^{2},...,p^{C})$
    \STATE \textbf{Output:} K value
    \STATE \textbf{Initialize:} cumulative probability upper bound $T$, category numbers $C$, cumulative probability $\bm{V}=\bm{p}$
    \STATE \textbf{Compute cumulative probability:} \\
    \FOR{$n=1$ \textbf{to} $C$}
        \IF{$V^{n} > T$ \textbf{or} $n = C$}
        \STATE return $n$
        \ENDIF
    \STATE $V^{n+1} = V^{n}+p^{n+1}$
    \ENDFOR
   \STATE \textbf{Determine K value:}\\
   \STATE $K=\max(n-1,1)$
   \STATE \textbf{Return} K
\end{algorithmic}
\end{algorithm}
Selecting top-n predicted categories whose cumulative probability exceeds $T$ guarantees that the ground truth has a high probability of being selected.
A counter-intuitive design in our Alg.~\ref{alg:K value selection strategy} is choosing $K=n-1$ instead of $K=n$.
This is because setting $K<n$ alleviates the gradient vanishing problem in training our FPL (cf. Appendix).
Another noteworthy point is that our algorithm provides $K=1$ for pixels with high confidence, while $K>1$ are usually supplied for uncertain pixels, as illustrated in Fig.~\ref{fig:k num pixel} and Fig.~\ref{fig:k map half}.
This property is in line with semantic intuition because a certain pixel should learn an explicit label, while an uncertain pixel needs to learn from multiple fuzzy labels.
The ablation study about the K value selection is in Appendix.

\subsection{Fuzzy Positive Regularization} \label{sec: pessimistic consistency loss}
In our FPA, we generate a fuzzy positive label set $\mathbb{Y}_{us} = \{y_{us}^{1},y_{us}^{2},...,y_{us}^{K}\}$ that contains K labels for each unlabeled pixel instead of only one pseudo label as in previous works.
Hence we need to propose a new loss function to learn the possible ground truth from $\mathbb{Y}_{us}$.

Our FPL regards all categories in the fuzzy positive label set $\mathbb{Y}_{us}$ are probable to be the ground truth, but the categories outside the $\mathbb{Y}_{us}$ are unlikely to be the ground truth.
Therefore, we hope that the predictions of our model for the $K$ fuzzy positive categories to be larger than the predictions for the rest $C-K$ negative categories.
We refer to some works in metric learning~\cite{liu2017sphereface,wang2018additive,wang2018cosface,sun2020circle} and formulate our optimization objective for each pixel as:
\begin{equation}
  \begin{aligned}
  \underset{i \in \mathbb{Y}_{us}} \min(z_{us}^{i})~~ >~~ \underset{j \notin \mathbb{Y}_{us}} \max(z_{us}^{j}),
\end{aligned} \label{eq:purpose}
\end{equation}
where $z_{us}^{i}$ represents the prediction of our model for the $i$-th category.
Eq.~\ref{eq:purpose} means we regularize the minimum of the predictions for categories in $\mathbb{Y}_{us}$ to be larger than the maximum of the predictions for other categories.
In other words, we enforce all the predictions for fuzzy positive categories to be larger than those for negative categories.
From Eq.~\ref{eq:purpose}, a straightforward loss function can be formulated as:
\begin{equation}
  \begin{aligned}
  \mathcal{L}^{f}_{us} = ReLU( \underset{j \notin \mathbb{Y}_{us}} \max(z_{us}^{j}) - \underset{i \in \mathbb{Y}_{us}} \min(z_{us}^{i}) ).
\end{aligned} \label{eq:loss function}
\end{equation}
However, this $\mathcal{L}_{us}^{f}$ is globally non-differentiable with respect to $\bm{z}_{us} = \{z_{us}^{1},z_{us}^{2},...,z_{us}^{C}\}$ since the $\max$ and $\min$ functions in Eq.~\ref{eq:loss function} are globally non-differentiable~\cite{pinter2001globally, mcelreath2018statistical}.
And the $ReLU$ function also has a singularity at $x=0$.
Thanks to existing functional approximations~\cite{nielsen2017guaranteed,mcelreath2018statistical,dugas2001incorporating,glorot2011deep}, we approximate the Eq.~\ref{eq:loss function} to make $\mathcal{L}_{us}^{f}$ differentiable:
\begin{equation}
  \begin{aligned}
  &\max (z^{1},z^{2},...,z^{n}) \approx \log(\sum_{i=1}^{n} \exp{(z^{i})}) \\
  &\min (z^{1},z^{2},...,z^{n}) \approx -\log(\sum_{i=1}^{n} \exp{(-z^{i})}) \\
  &ReLU(z) = \max(z,0) \approx \log(1+\exp(z)).
\end{aligned} \label{eq:functional approximation}
\end{equation}
Based on these functional approximations, our fuzzy positive consistency loss $\mathcal{L}^{f}$ for one pixel $x_{us}$ (i.e., the $s$-th pixels of the $u$-th unlabeled image) could be converted to:
\begin{equation}
  \begin{aligned}
  \mathcal{L}_{us}^{f} = \log(1+\sum_{i \in \mathbb{Y}_{us}} e^{-z_{us}^{i}} \times \sum_{j \notin \mathbb{Y}_{us}} e^{z_{us}^{j}}).
\end{aligned} \label{eq:topk consistency loss}
\end{equation}

Next, we analyze the behavior of $\mathcal{L}_{us}^{f}$ in backpropagation.
The gradient of $\mathcal{L}_{us}^{f}$ with respect to the prediction $\bm{z}_{us}$ of our model is computed as:
\begin{equation}
  \begin{aligned}
    &\frac{\partial \mathcal{L}_{us}^{f}}{\partial z_{us}^{i}} = \frac{-\sum_{j \notin \mathbb{Y}_{us}}{e^{z_{us}^{j}}} \times e^{-z_{us}^{i}}}{1+\sum_{j \notin \mathbb{Y}_{us}}{e^{z_{us}^{j}}} \times \sum_{i \in \mathbb{Y}_{us}}{e^{-z_{us}^{i}}}},i\in \mathbb{Y}_{us}\\
    &\frac{\partial \mathcal{L}_{us}^{f}}{\partial z_{us}^{j}} = \frac{\sum_{i \in \mathbb{Y}_{us}}{e^{-z_{us}^{i}}} \times e^{z_{us}^{j}}}{1+\sum_{j \notin \mathbb{Y}_{us}}{e^{z_{us}^{j}}} \times \sum_{i \in \mathbb{Y}_{us}}{e^{-z_{us}^{i}}}},j\notin \mathbb{Y}_{us},
  \end{aligned} \label{eq:topk_gradient}
\end{equation}
where the $\frac{\partial \mathcal{L}_{us}^{f}}{\partial z_{us}^{i}}$ and $\frac{\partial \mathcal{L}_{us}^{f}}{\partial z_{us}^{j}}$ denote the derivatives with respect to predictions for fuzzy positive categories and other negative categories, respectively.
From Eq.~\ref{eq:topk consistency loss} and Eq.~\ref{eq:topk_gradient}, we see that our $\mathcal{L}_{us}^{f}$ has following characteristics:

1) The prediction for the ground truth increases when it appears in $\mathbb{Y}_{us}$.
This is because predictions for fuzzy positive categories have gradients less than 0, and thus are optimized to increase by gradient descent.

2) The existing $\mathcal{L}_{us}^{v}$ is a special case of our $\mathcal{L}_{us}^{f}$ when we set $K=1$, as shown in Eq.~\ref{eq:special case}.
\begin{equation}
  \begin{aligned}
    \mathcal{L}_{us}^{v} = \log(1+e^{-z_{us}^{i}} \times \sum_{j \neq i}e^{z_{us}^{j}}),
  \end{aligned} \label{eq:special case}
\end{equation}
where $i$ is the index of the top-1 predicted pseudo label.

\begin{figure*}[htbp]
\centerline{\includegraphics[scale=0.25]{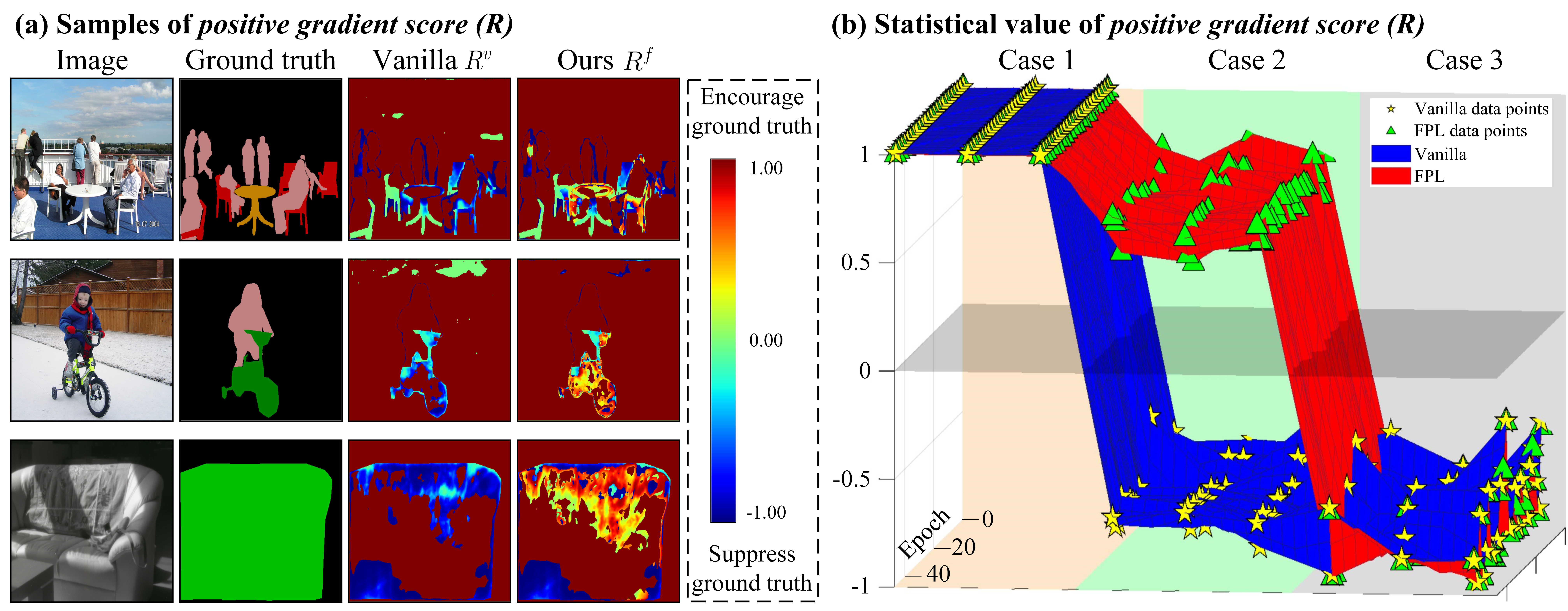}}
\caption{
\textbf{Positive gradient score $\bm{R}$.}
(a) shows the positive gradient score maps of some unlabeled examples, where the red color means the prediction of ground truth is encouraged, while the blue color indicates suppression.
(b) is the statistics value of the positive gradient score in three cases (Sec.~\ref{sec:gradient ratio}).
This figure is plotted on VOC2012 with 1/16 labeled data.
}
\label{fig:gradient_ratio}
\vspace{-1.0em}
\end{figure*}

\textbf{Adaptive weight for each pixel:}\label{sec:adaptive weight}
From Eq.~\ref{eq:purpose}, it can be seen that our model learns informative semantics based on the assumption that the ground truth exists in the fuzzy positive label set $\mathbb{Y}_{us}$.
Thus, we propose to integrate the confidence of this assumption into the training of FPL.
When our assumption is not tenable, the ground truth will be outside $\mathbb{Y}_{us}$, and its largest predicted probability is $\underset{j \notin \mathbb{Y}_{us}} \max(p_{us}^{j})$.
Therefore, the $\underset{j \notin \mathbb{Y}_{us}} \max(p_{us}^{j})$ is negatively correlated with the assumption confidence since high $\underset{j \notin \mathbb{Y}_{us}} \max(p_{us}^{j})$ means ground truth has a low probability inside $\mathbb{Y}_{us}$, and vice versa.

Formulately, the range of $\underset{j \notin \mathbb{Y}_{us}} \max(p_{us}^{j})$ is derived as:
\begin{equation}
  \begin{aligned}
    \frac{1-T}{C-K_{us}}<\underset{j \notin \mathbb{Y}_{us}} \max(p_{us}^{j}) < \frac{\sum_{i \in \mathbb{Y}_{us}}{p_{us}^{i}}}{K_{us}}.
  \end{aligned} \label{eq:constraint2}
\end{equation}
In practice, $T$ is close to 1 (e.g., 0.9), thus $\frac{1-T}{C-K_{us}}$ is close to 0.
For simplicity, we obtain the approximate range of $\underset{j \notin \mathbb{Y}_{us}} \max(p_{us}^{j})$ as $(0, \frac{\sum_{i \in \mathbb{Y}_{us}}{p_{us}^{i}}}{K_{us}})$.
We then define our adaptive weight as a monotonically decreasing concave function:
\begin{equation}
  \begin{aligned}
    w_{us} = \frac{\log{[1+A \times (\frac{\sum_{i \in \mathbb{Y}_{us}}{p_{us}^{i}}}{K_{us}}-\underset{j \notin \mathbb{Y}_{us}} \max(p_{us}^{j}))}]}{\log{[1+A \times (\frac{\sum_{i \in \mathbb{Y}_{us}}{p_{us}^{i}}}{K_{us}})]}},
  \end{aligned} \label{eq:weight}
\end{equation}
where $A$ is a scalar used to control the radian of this function, which is fixed as 50.
It is worth noting that our adaptive weight is different from the weights computed by top-1 confidence used to filter out or re-weight low-confidence pixels~\cite{french2019semi,ouali2020semi,ke2020guided}.
Those weights are small for pixels with low top-1 probability, resulting in those pixels not being sufficiently used in training~\cite{wang2022semi}.
But our weight is only small when the prediction of a pixel is confused in the top-(K+1) categories, thus our model still uses the information that its prediction should not belong to other C-K-1 categories.

\subsection{Principle Analysis}\label{sec:gradient ratio}


Ideally, we hope to learn the semantics of ground truth in unlabeled data, but in practice, we can only learn the semantics of positive categories and suppress the rest.
Here, we propose a \textit{positive gradient score} $R$ to measure how properly the ground truth is learned :
\begin{equation}
  \begin{aligned}
  R_{us} = \frac{\partial \mathcal{L}_{us}}{\partial z_{us}^{gt}} ~~/~~ \sum_{i \in Y_{us}} \frac{\partial \mathcal{L}_{us}}{\partial z_{us}^{i}},
  \end{aligned} \label{eq:ce_gradient}
\end{equation}
where the $Y_{us}$ represents the fuzzy positive label set $\mathbb{Y}_{us}$ when $\mathcal{L}_{us}$ is $\mathcal{L}_{us}^{f}$, and $Y_{us}$ represents the pseudo label when $\mathcal{L}_{us}$ is $\mathcal{L}_{us}^{v}$.
The positive gradient score $R_{us}$ is the ratio of the gradient for the ground truth to the sum of the gradients for all positive categories.
It ranges from $[-1,1]$ and a positive $R_{us}$ means the GT prediction is encouraged to increase, while a negative $R_{us}$ means the GT prediction is incorrectly suppressed to decrease.
Based on actual training, we consider $R_{us}$ in three cases:

\textbf{Case 1}. The pseudo label is correct, that is, the ground truth is the top-1 predicted category.
In this case, the positive gradient score $R_{us}$ computed by $\mathcal{L}_{us}^{v}$ and $\mathcal{L}_{us}^{f}$ are:
\begin{equation}
  \begin{aligned}
    R_{us}^{v} = \frac{p_{us}^{gt}-1}{p_{us}^{pse}-1}=1,~~R_{us}^{f} = \frac{e^{-z_{us}^{gt}}}{\sum_{i \in \mathbb{Y}_{us}} e^{-z_{us}^{i}}} \in [0,1],
  \end{aligned} \label{eq:case1_score_ratio}
\end{equation}
where $p_{us}^{gt}$ and $p_{us}^{pse}$ are the predicted probabilities for ground truth and the pseudo-label category.
When the size of $\mathbb{Y}_{us}$ (i.e., K value) is $1$, the $R_{us}^{f}$ will be equal to $R_{us}^{v}$ as $1$.
We see that $R_{us}^{f}$ and $R_{us}^{v}$ are both greater than 0, meaning they both encourage the GT prediction to increase.
In practice, the statistics of $R_{us}^{f}$ is close to $1$.
This is because most pixels in this case have $K=1$ (cf. Appendix).


\textbf{Case 2}. The top-1 prediction is wrong, but the ground truth is in the categories with top-K probabilities, where K is computed by our K value selection strategy in Alg.~\ref{alg:K value selection strategy}.
For Case 2, the positive gradient score $R_{us}^{v}$ and $R_{us}^{f}$ are computed as:
\begin{equation}
  \begin{aligned}
    R_{us}^{v} = \frac{p_{us}^{gt}}{p_{us}^{pse}-1} \in [-1,0],~~R_{us}^{f} = \frac{e^{-z_{us}^{gt}}}{\sum_{i \in \mathbb{Y}_{us}} e^{-z_{us}^{i}}} \in [0,1].
  \end{aligned} \label{eq:case2_score_ratio}
\end{equation}
We see that $R_{us}^{f}$ is larger than 0 while $R_{us}^{v}$ is less than 0.
This is because the ground truth is missed by the pseudo label but captured by our fuzzy positive label set.
It means that vanilla $\mathcal{L}_{us}^{v}$ erroneously suppresses GT prediction, but our $\mathcal{L}_{us}^{f}$ encourages GT prediction, reflecting FPL remarkably reduces the interference from wrong pseudo labels.

\textbf{Case 3}. The pseudo label is wrong, and the ground truth is also outside the fuzzy positive labels $\mathbb{Y}_{us}$.
In this case, the positive gradient score $R^{v}_{us}$ and $R^{f}_{us}$ are:
\begin{equation}
  \begin{aligned}
    R_{us}^{v} = \frac{p_{us}^{gt}}{p_{us}^{pse}-1} \in [-1,0],~~R_{us}^{f} = \frac{-e^{z_{us}^{gt}}}{\sum_{j \notin \mathbb{Y}_{us}} e^{z_{us}^{j}}} \in [-1,0].
  \end{aligned} \label{eq:case3_score_ratio}
\end{equation}
It is obvious that $R_{us}^{f}$ and $R_{us}^{v}$ are both less than 0, meaning neither $\mathcal{L}_{us}^{v}$ nor $\mathcal{L}_{us}^{f}$ is beneficial for learning the semantics of ground truth in this case.
In Fig.~\ref{fig:gradient_ratio} (a), we display some examples which intuitively reflect the advantages of $R_{us}^{f}$ over $R_{us}^{v}$.
That is, many parts of $R_{us}^{v}$ less than 0 (colored in blue) becomes larger than 0 in $R_{us}^{f}$ (colored in red).
In Fig.~\ref{fig:gradient_ratio} (b), the statistics of positive gradient score show $R_{us}^{f}$ significantly outperforms the existing $R_{us}^{v}$ in Case 2, and they perform similarly in Case 1 and Case 3.

\begin{table*}[t]\footnotesize
  \setlength{\tabcolsep}{1.8mm}
  \centering
  \begin{tabular}{lcccccccc}
  \hline
  \multicolumn{1}{l|}{\multirow{2}{*}{Method}} & \multicolumn{4}{c|}{ResNet 50}                                                 & \multicolumn{4}{c}{ResNet 101}                            \\ \cline{2-9}
  \multicolumn{1}{l|}{}                        & 1/32 (93)        & 1/16 (186)        & 1/8 (372)          & \multicolumn{1}{c|}{1/4 (744)}          & 1/32 (93)        & 1/16 (186)        & 1/8 (372)         & 1/4 (744)         \\ \hline
  \multicolumn{1}{l|}{MT~\cite{tarvainen2017mean}}          & -        & 66.14        & 72.03        & \multicolumn{1}{c|}{74.47}        & -        & 68.08        & 73.71        & 76.53        \\
  \multicolumn{1}{l|}{CCT~\cite{ouali2020semi}}          & -        & 66.35        & 72.46        & \multicolumn{1}{c|}{75.68}        & -        & 69.64        & 74.48        & 76.35        \\
  \multicolumn{1}{l|}{GCT~\cite{ke2020guided}}          & -        & 65.81        & 71.33        & \multicolumn{1}{c|}{75.30}        & -        & 66.90        & 72.96        & 76.45        \\
  \multicolumn{1}{l|}{U$^{2}$PL~\cite{wang2022semi}}                    & -        & -        & -        & \multicolumn{1}{c|}{-}        & -        & 74.90        & 76.48        & 78.51        \\ \hline

  \rowcolor{orange!5}\multicolumn{1}{l|}{CPS w/o cutmix$^{\dag}$~\cite{chen2021semi}}          & 54.40        & 68.68        & 73.06        & \multicolumn{1}{c|}{75.75}        & 59.70        & 71.22        & 74.98        & 77.45        \\
  \rowcolor{orange!5}\multicolumn{1}{l|}{\textbf{FPL}+CPS w/o cutmix}      & 55.77(\textcolor{red}{$\uparrow$1.37}) & 69.71(\textcolor{red}{$\uparrow$1.03}) & 74.43(\textcolor{red}{$\uparrow$1.37}) & \multicolumn{1}{c|}{76.76(\textcolor{red}{$\uparrow$1.01})} & 61.00(\textcolor{red}{$\uparrow$1.30})        & 72.05(\textcolor{red}{$\uparrow$0.83})        & 75.67(\textcolor{red}{$\uparrow$0.69})        & 77.57(\textcolor{red}{$\uparrow$0.12})        \\\hline
  \rowcolor{orange!5}\multicolumn{1}{l|}{CPS w/ cutmix$^{\dag}$~\cite{chen2021semi}}           & 71.33        & 74.05        & 76.92        & \multicolumn{1}{c|}{77.77}        & 72.51        & 74.72        & 77.62        & 78.93        \\
  \rowcolor{orange!5}\multicolumn{1}{l|}{\textbf{FPL}+CPS w/ cutmix}       & \textbf{72.39}(\textcolor{red}{$\uparrow$1.06})        & \textbf{74.80}(\textcolor{red}{$\uparrow$0.75})        & \textbf{77.32}(\textcolor{red}{$\uparrow$0.40})        & \multicolumn{1}{c|}{\textbf{78.53}(\textcolor{red}{$\uparrow$0.76})}        & 73.20(\textcolor{red}{$\uparrow$0.69}) & 75.74(\textcolor{red}{$\uparrow$1.02}) & \textbf{78.47}(\textcolor{red}{$\uparrow$0.85}) & \textbf{79.19}(\textcolor{red}{$\uparrow$0.26}) \\\hline
  \rowcolor{orange!5}\multicolumn{1}{l|}{AEL$^{\dag}$~\cite{hu2021semi}}                     & 68.39        & 74.03        & 75.83        & \multicolumn{1}{c|}{76.18}        & 73.00        & 75.26        & 78.07        & 78.26        \\
  \rowcolor{orange!5}\multicolumn{1}{l|}{\textbf{FPL}+AEL}                 & 71.21(\textcolor{red}{$\uparrow$2.82})        & 74.54(\textcolor{red}{$\uparrow$0.51})        & 76.25(\textcolor{red}{$\uparrow$0.42})        & \multicolumn{1}{c|}{76.88(\textcolor{red}{$\uparrow$0.70})}        & \textbf{75.01}(\textcolor{red}{$\uparrow$2.01}) & \textbf{76.58}(\textcolor{red}{$\uparrow$1.32}) & 78.19(\textcolor{red}{$\uparrow$0.12}) & 78.46(\textcolor{red}{$\uparrow$0.20})        \\ \hline
  \end{tabular}
  \vspace{-0.5em}
  \caption{\textbf{The mIoU on Cityscapes.} Results marked by $\dag$ are reproduced in the same experimental environment as FPL.}\label{tab:cityscapes results}
\end{table*}

\begin{table*}[t]\footnotesize
  \setlength{\tabcolsep}{4.4mm}
  \centering
  \begin{tabular}{l|ccc|ccc}
\hline
\multirow{2}{*}{Method} & \multicolumn{3}{c|}{ResNet 50}             & \multicolumn{3}{c}{ResNet 101}             \\ \cline{2-7}
                        & 1/16 (662)         & 1/8 (1323)          & 1/4 (2646)          & 1/16 (662)      & 1/8 (1323)          & 1/4 (2646)          \\ \hline
MT~\cite{tarvainen2017mean}                      & 66.77        & 70.78        & 73.22        & 70.59        & 73.20        & 76.62        \\
CCT~\cite{ouali2020semi}                     & 65.22        & 70.87        & 73.43        & 67.94        & 73.00        & 76.17        \\
CutMix-Seg~\cite{french2019semi}              & 68.90        & 70.70        & 72.46        & 72.56        & 72.69        & 74.25        \\
GCT~\cite{ke2020guided}                     & 64.05        & 70.47        & 73.45        & 69.77        & 73.30        & 75.25        \\
CAC~\cite{lai2021semi}                     & 70.10        & 72.40        & 74.00        & 72.40        & 74.60        & 76.30        \\ \hline
\rowcolor{orange!5}CPS w/o cutmix$^{\dag}$~\cite{chen2021semi}          & 68.13        & 72.79        & 74.24        & 72.50        & 74.97        & 77.14        \\
\rowcolor{orange!5}\textbf{FPL}+CPS w/o cutmix      & 68.67(\textcolor{red}{$\uparrow$0.54}) & 73.03(\textcolor{red}{$\uparrow$0.36}) & 74.80(\textcolor{red}{$\uparrow$0.56}) & 73.18(\textcolor{red}{$\uparrow$0.68}) & 75.74(\textcolor{red}{$\uparrow$0.77}) & 77.47(\textcolor{red}{$\uparrow$0.33}) \\\hline
\rowcolor{orange!5}CPS w/ cutmix$^{\dag}$~\cite{chen2021semi}           & 71.78        & 73.44        & 74.90        & 74.48        & 76.44        & 77.68        \\
\rowcolor{orange!5}\textbf{FPL}+CPS w/ cutmix       & \textbf{72.52}(\textcolor{red}{$\uparrow$0.74}) & \textbf{73.74}(\textcolor{red}{$\uparrow$0.30}) & 75.35(\textcolor{red}{$\uparrow$0.45}) & 74.98(\textcolor{red}{$\uparrow$0.50}) & \textbf{77.75}(\textcolor{red}{$\uparrow$1.31}) & 78.30(\textcolor{red}{$\uparrow$0.62}) \\\hline
\rowcolor{orange!5}AEL$^{\dag}$~\cite{hu2021semi}                     & 69.93        & 73.17        & 75.50        & 74.20        & 76.58        & 77.98        \\
\rowcolor{orange!5}\textbf{FPL}+AEL                 & 71.01(\textcolor{red}{$\uparrow$1.08})        & 73.69(\textcolor{red}{$\uparrow$0.52})        & \textbf{76.61}(\textcolor{red}{$\uparrow$1.11})        & \textbf{74.98}(\textcolor{red}{$\uparrow$0.78}) & 76.73(\textcolor{red}{$\uparrow$0.15}) & \textbf{78.35}(\textcolor{red}{$\uparrow$0.37}) \\ \hline
\end{tabular}
\vspace{-0.5em}
\caption{\textbf{The mIoU on VOC2012.} Results marked by $\dag$ are reproduced in the same experimental environment as FPL.}\label{tab:voc standard results}
\vspace{-1.0em}
\end{table*}

\begin{table}[t]\footnotesize
  \setlength{\tabcolsep}{0.4mm}
\begin{tabular}{lcccc}
\hline
\multicolumn{1}{l|}{Method}              & 1/16 (92)        & 1/8 (183)         & 1/4 (366)         & 1/2 (732)       \\ \hline
\multicolumn{1}{l|}{AdvSemSeg~\cite{hung2018adversarial}}              & 39.69        & 47.58        & 59.97        & 65.27        \\
\multicolumn{1}{l|}{CCT~\cite{ouali2020semi}}              & 33.10        & 47.60        & 58.80        & 62.10        \\
\multicolumn{1}{l|}{VAT~\cite{miyato2018virtual}}              & 36.92        & 49.35        & 56.88        & 63.34        \\
\multicolumn{1}{l|}{MT~\cite{tarvainen2017mean}}              & 48.70        & 55.81        & 63.01        & 69.16        \\
\multicolumn{1}{l|}{GCT~\cite{ke2020guided}}              & 46.04        & 54.98        & 64.71        & 70.67        \\
\multicolumn{1}{l|}{CutMix-Seg~\cite{french2019semi}}              & 52.16        & 63.47        & 69.46        & 73.73        \\
\multicolumn{1}{l|}{PseusoSeg~\cite{zou2020pseudoseg}}               & 57.60        & 65.50        & 69.14        & 72.41        \\
\multicolumn{1}{l|}{PC$^{2}$Seg~\cite{zhong2021pixel}}                    & 57.00        & 66.28        & 69.78        & 73.05        \\
\multicolumn{1}{l|}{U$^{2}$PL~\cite{wang2022semi}}                    & 67.98        & 69.15        & 73.66        & 76.16        \\ \hline
\rowcolor{orange!5}\multicolumn{1}{l|}{CPS w/ cm$^{\dag}$~\cite{chen2021semi}}           & 67.53        & 70.41        & 75.27        & 78.69        \\
\rowcolor{orange!5}\multicolumn{1}{l|}{\textbf{FPL}+CPS w/ cm}       & \textbf{69.30}(\textcolor{red}{$\uparrow$1.77}) & \textbf{71.72}(\textcolor{red}{$\uparrow$1.31}) & \textbf{75.73}(\textcolor{red}{$\uparrow$0.46}) & \textbf{78.95}(\textcolor{red}{$\uparrow$0.26}) \\ \hline
\end{tabular}
\vspace{-0.5em}
\caption{\textbf{The mIoU on VOC2012 LowData}. Results marked by $\dag$ are reproduced in the same experimental environment as FPL. The `cm' is the cutmix.}\label{tab:voc lowdata results}
\end{table}


\section{Experiments}
\subsection{Implementation Details}
\textbf{Frameworks and dataset:}
We evaluate the effectiveness of our FPL on two widely used frameworks, CPS~\cite{chen2021semi} and AEL~\cite{hu2021semi}, and two datasets \textit{PASCAL VOC 2012} and \textit{Cityscapes}.
The Cityscapes is a large-scale dataset designed for urban street scene segmentation which consists of 19 semantic classes containing 2,975 images for training, 500 for validation, and 1,525 for testing.
The PASCAL VOC 2012 is a generic object segmentation benchmark that consists of 20 object classes and 1 background class.
It is divided into training, validation, and test sets including 1,464, 1,449, and 1,456 images, respectively.
There is also an augmented set~\cite{hariharan2011semantic} adding 10,582 images into the standard training set.
Following the setting of previous works~\cite{zou2020pseudoseg,chen2021semi}, we implement two splits on VOC2012: standard split (with augmented set) and low data split (without augmented set).

\textbf{Experimental setting:}
Following the default settings of CPS and AEL, we use Deeplab v3+ with pre-trained ResNet-50 and ResNet-101 as backbones.
Specifically, on Cityscapes using CPS as the baseline, we use SGD optimizer with a weight decay of 1e-4.
The initial learning rate is set to 0.02 and the momentum is fixed at 0.9.
We use the default ‘poly’ learning rate decay policy to scale the learning rate by $(1-iter/max~iter)^{0.9}$, and this policy is used in all our experiments.
The input images are cropped to $800 \times 800$ and the batchsize is 64.
When using AEL as the baseline, the batchsize, learning rate, and image size are changed to 16, 0.01, and $769 \time 769$.
On VOC2012 using CPS as the baseline, we use SGD optimizer with a weight decay of 1e-4.
The initial learning rate is set to 0.01 and the momentum is fixed at 0.9.
The input images are cropped to $512 \times 512$ and the batchsize is 32.
When using AEL as the baseline, the batchsize is changed to 16.
The cumulative probability upper bound $T$ in all our experiments is set from \{0.95, 0.9, 0.85\}. More details are in Appendix.

\subsection{Quantitative Results}
Our FPL model is trained with the same hyperparameters as the baseline model, only replacing the vanilla positive learning using one pseudo label with our fuzzy positive learning using multiple fuzzy positive labels.
The segmentation results on Cityscapes, VOC2012, and VOC2012 LowData are presented in Table~\ref{tab:cityscapes results}, Table~\ref{tab:voc standard results}, and Table~\ref{tab:voc lowdata results}, where red numbers represent the improvement brought by FPL to the baseline.
We see that FPL achieves stable improvements over baseline models across all data splits.
Besides, FPL improves the CPS baseline under both with and without CutMix settings, indicating that the performance gain from FPL and data augmentation (e.g., CutMix) can be accumulated.
Furthermore, FPL is effective on multiple baselines, i.e., CPS and AEL, which means FPL is universal for various existing SSL frameworks.

\subsection{Empirical Study}

\subsubsection{The Hyperparameter $T$}
The $T$ is the only new hyperparameter brought by FPL, which controls the K values of pixels in training.
Here we summarize two rules for setting a proper $T$ value.
First, a $T$ value around 0.9 (e.g., 0.85, 0.9, 0.95) is usually a promising setting.
Second, a $T$ value set negatively correlated to the number of labeled data usually brings high performance.

\textbf{The effect of $T$ on the training behaviors.}
In training, $T$ affects the number of fuzzy positive labels for each pixel (K value), which reflects the degree of fuzziness of our FPL. Besides, $T$ also affects the impurity of the fuzzy positive label set, which is the proportion of pixels whose ground truths are missed in fuzzy positive labels.
We formulate the average K value and the impurity as:
\begin{equation}
  \centering
  \begin{aligned}
    \overline{K} &= \frac{1}{U \times S} \times \sum_{u=1}^{U} \sum_{s=1}^{S} K_{us} \\
    impurity &= \frac{1}{U\times S} \times \sum_{u=1}^{U} \sum_{s=1}^{S} \mathbbm{1}(y_{us} \notin \mathbb{Y}_{us}),
  \end{aligned} \label{eq:avg K and purity}
\end{equation}
where $K_{us}$, $y_{us}$, and $\mathbb{Y}_{us}$ are the K value, ground truth, and fuzzy positive label set for the $s$-th pixel of the $u$-th image.
The $U$ is the size of the unlabeled dataset, and $S$ is the number of pixels contained in each image.

\begin{figure}[t]
\centerline{\includegraphics[scale=0.35]{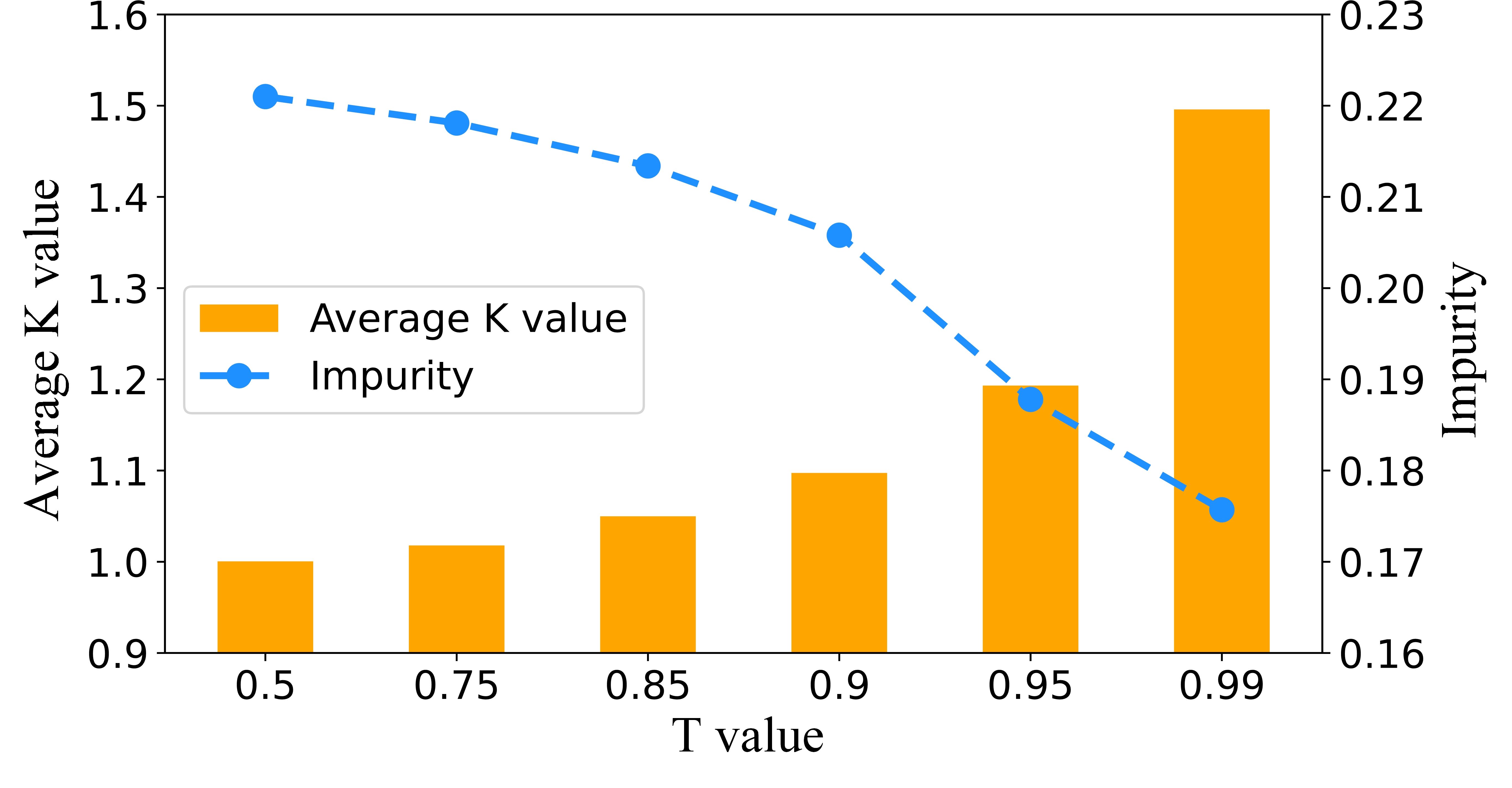}}
\vspace{-1.0em}
\caption{
\textbf{The ablation study about our cumulative probability upper bound $T$.} The results are obtained on Cityscapes with 1/16 labeled data using CPS as the baseline.}
\label{tab:cumulative probability upperbound T}
\vspace{-1.0em}
\end{figure}

In Fig.~\ref{tab:cumulative probability upperbound T}, we provide the average K value and impurity.
We see that small $T$ values lead to small K values, meaning the size of the fuzzy positive label set is small.
Accordingly, it causes high impurity since a small fuzzy positive label set has a relatively large possibility of missing the ground truth.
High impurity misleads the learning process, which is known as confirmation bias~\cite{arazo2020pseudo}.
In contrast, using a large value of $T$ builds a large fuzzy positive label set, effectively reducing the impurity.
However, too large $T$ value (e.g. 0.99) makes our model learn from too many labels, which is also not suitable for a single-label classification task.
In Table~\ref{tab:cumulative probability upperbound T mIoU}, we present the mIoU of our FPL models trained with various $T$ values.
Given our observation on the trade-off between the impurity and the size of the fuzzy positive label set, we find a $T$ value around 0.9 always provides promising results.

\begin{table}[t]\footnotesize
\setlength{\tabcolsep}{2.6mm}
\begin{tabular}{l|cccccc}
\hline
T value & 0.5   & 0.75  & 0.85  & 0.9   & 0.95  & 0.99  \\ \hline
mIoU    & 68.80 & 68.97 & \textcolor{red}{69.34} & \textbf{\textcolor{red}{69.71}} & \textcolor{red}{69.08} & 67.52 \\ \hline
\end{tabular}
\vspace{-0.5em}
\caption{\textbf{The performances of FPL models with various $T$.} These results are obtained on Cityscapes with 1/16 labeled data using CPS as the baseline.}\label{tab:cumulative probability upperbound T mIoU}
\end{table}


\begin{table}[t]\footnotesize
\setlength{\tabcolsep}{4.0mm}
\centering
\begin{tabular}{l|ccc}
\hline
T    & 1/32          & 1/16          & 1/8           \\ \hline
0.85 & 55.22 (\textcolor{red}{$\uparrow$0.90}) & 69.34 (\textcolor{red}{$\uparrow$0.66}) & 74.37 (\textcolor{red}{$\uparrow$1.31}) \\
0.9  & 55.40 (\textcolor{red}{$\uparrow$1.08}) & 69.71 (\textbf{\textcolor{red}{$\uparrow$1.03}}) & 74.43 (\textbf{\textcolor{red}{$\uparrow$1.37}}) \\
0.95 & 55.77 (\textbf{\textcolor{red}{$\uparrow$1.45}}) & 69.08 (\textcolor{red}{$\uparrow$0.40}) & 74.03 (\textcolor{red}{$\uparrow$0.97}) \\ \hline
\end{tabular}
\vspace{-0.5em}
\caption{\textbf{The relationship between cumulative probability upper bound $T$ and the amount of labeled data.} The results are obtained on Cityscapes using CPS as the baseline.}\label{tab:cumulative probability upperbound labeled amount}
\end{table}

\begin{figure}[t]
\centerline{\includegraphics[scale=0.46]{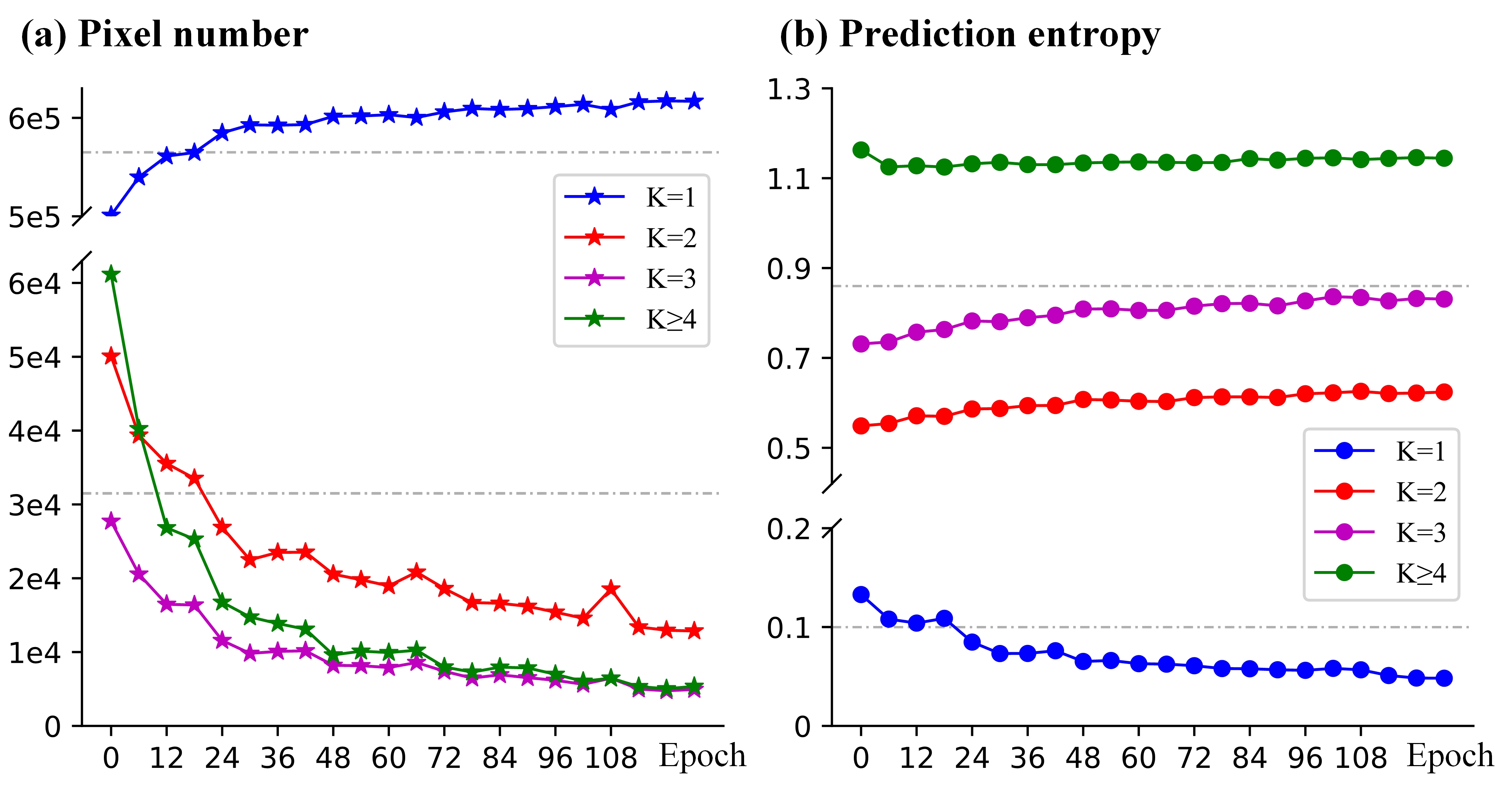}}
\caption{
(a) The K values during training. (b) The prediction entropy of pixels with various K values.
The input size is $800 \times 800$ meaning there are 640000 pixels in total.
}
\label{fig:k num pixel}
\vspace{-1.0em}
\end{figure}

\textbf{The relationship between $T$ and the amount of labeled data.}
We find a high $T$ usually obtains good performance when labeled data are limited, while a low $T$ usually performs better when labeled data are sufficient.
As shown in Table~\ref{tab:cumulative probability upperbound labeled amount}, in the 1/32 labeled data setting, $T=0.95$ obtains the highest improvement about 1.45\%, while $T=0.85$ and $T=0.9$ only obtain improvements about 1\%.
In the 1/16 labeled data setting, $T=0.9$ obtains the best performance, improving baseline by 1.03\%, and the rest two $T$ values improve baseline by about 0.5\%.
In the 1/8 labeled data setting, $T=0.9$ and $T=0.85$ obtains close performances which improve baseline by 1.3\%, while $T=0.95$ performs not as well as the previous two $T$ settings.
It is obvious that setting the $T$ value negatively according to the amount of labeled data significantly benefits the performance.

\subsubsection{K Values in Training}
The number of pixels with different $K$ values is shown in Fig.~\ref{fig:k num pixel} (a).
We see that within training, the number of pixels with $K > 1$ decreases and the number of pixels with $K = 1$ increases.
At the late stage of training, the K values for more than 93.75\% (i.e., 6e5~/~6.4e5) pixels are 1.
This indicates the K values automatically converge to 1, meaning FPL could progressively achieve clear pixel-level semantic discrimination.
In Fig.~\ref{fig:k num pixel} (b), we illustrate that our FPL provides $K=1$ for certain pixels with low entropy while providing $K>1$ for uncertain pixels with high entropy.


\begin{figure}[t]
\centerline{\includegraphics[scale=0.29]{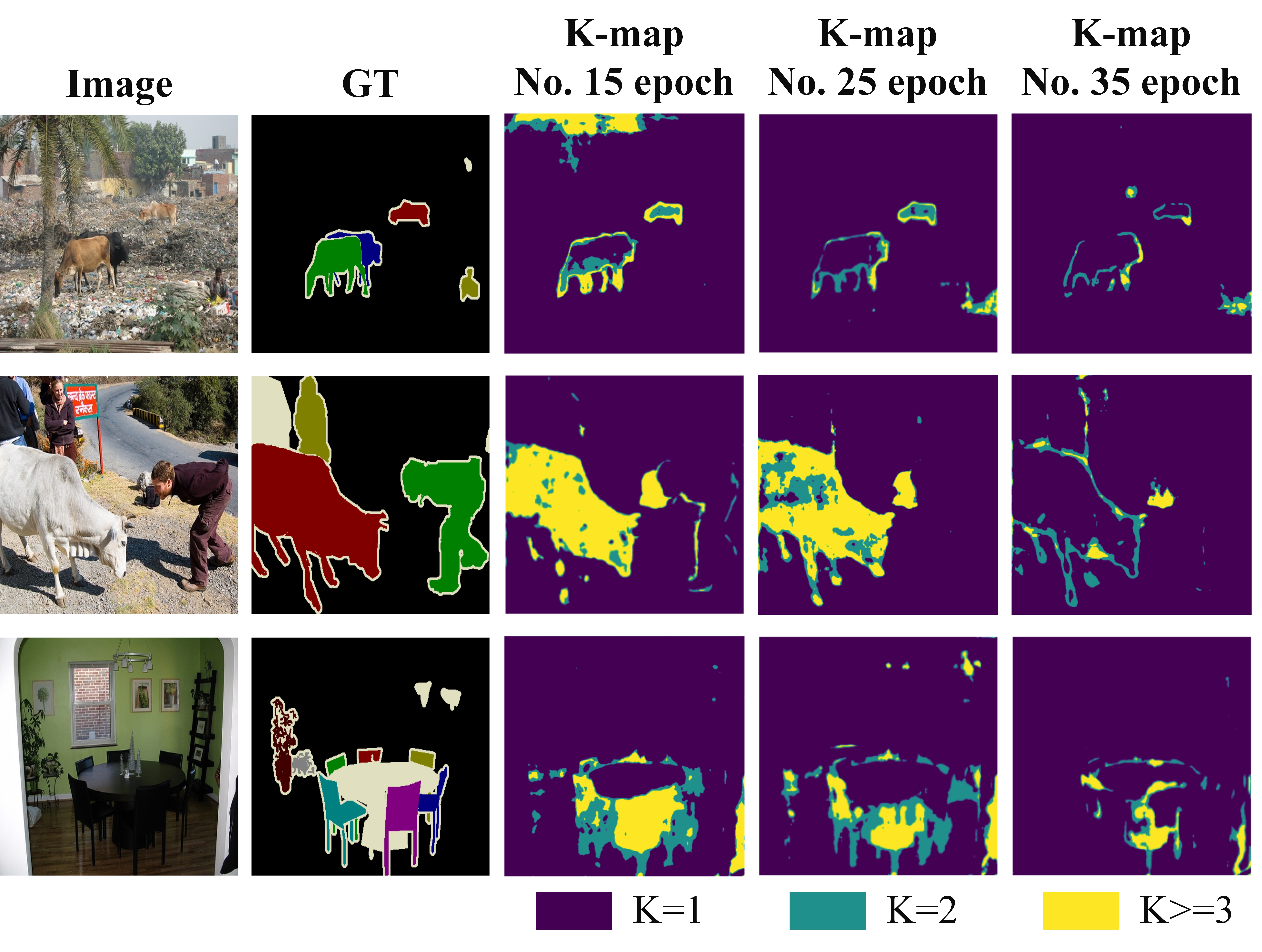}}
\vspace{-0.5em}
\caption{
\textbf{K value visualization,} plotted on the VOC2012 with 1/16 labeled data using FPL+CPS w/ CutMix.
}
\label{fig:k map half}
\vspace{-1.5em}
\end{figure}

Moreover, we show the K value maps of some examples during training in Fig.~\ref{fig:k map half}.
We see that the K values of most pixels in the background are 1 since background pixels are usually easy to classify.
In the early stage of training, the pixels with $K > 1$ are mainly located on objects, since the classification of objects for our model is uncertain at early training.
As the training progresses, the number of pixels with $K > 1$ gradually decreases and these pixels are mainly located at the boundary of objects.
This is because our model has certain predictions for most pixels in the later stage of training.
But for pixels located at the object boundary, their categories are fuzzy, for which our model makes uncertain predictions for them.
Our FPL provides multiple labels (i.e., $K>1$) for these uncertain pixels to learn, which is in line with their fuzzy property.

\section{Conclusion}
In this paper, we introduce a novel plug-and-play method named FPL for semi-supervised semantic segmentation.
Our method is the first to explore learning the semantics of ground truth from multiple fuzzy positive labels.
Specifically, We first propose a fuzzy positive assignment algorithm to provide an adaptive number of labels for each pixel.
We then develop a fuzzy positive regularization to learn the possible ground truth from these fuzzy positive labels.
Extensive experiments on two commonly used benchmarks with consistent performance gain demonstrate the effectiveness of our method.
Moreover, we provide an analysis showing the superiority of FPL in that it revises the gradient of learning ground truth when pseudo labels are wrong.
There are still directions worth continuing to explore in FPL,
e.g., ``extending discrete K values to continuous form for finer-grained fuzzy positive labels."
{\small
\bibliographystyle{ieee_fullname}
\bibliography{egbib}
}

\appendixpage
\setcounter{equation}{0}
\setcounter{section}{0}
\setcounter{figure}{0}
\setcounter{table}{0}
\section{Overview}
In this appendix, we provide the detailed setting about $T$ in Sec.~\ref{sec:T}.
For more analyses about our FPL, difference between FPL and negative learning is in Sec.~\ref{sec:neg}, gradient vanishing in K value selection strategy is in Sec.~\ref{sec: gradient vanishing}, details of positive gradient score is in Sec.~\ref{sec:score}, and gradient similarity between $G^{f}$ and $G^{ide}$ is in Sec.~\ref{sec:cos_sim}.
For more ablation studies about adaptive weight and K value selection strategy are in Sec.\ref{sec: adaptive weight} and Sec.~\ref{sec: K strategy}, respectively.
Besides, we discuss the limitation of our FPL in Sec.\ref{sec:limitation} and illustrate more examples in Sec.~\ref{sec:vis}.

\section{Experimental Details}\label{sec:T}
We provide the detailed setting about the cumulative probability upper bound $T$ in our experiments in Table~\ref{T city}, Table~\ref{T voc}, Table~\ref{T voc lowdata}.
\begin{table}[htbp]\footnotesize
\setlength{\tabcolsep}{1.0mm}
  \centering
  \caption{\textbf{The setting of $T$ on Cityscapes.}}
\begin{tabular}{l|cccc|cccc}
\hline
\multirow{2}{*}{Method} & \multicolumn{4}{c|}{ResNet 50} & \multicolumn{4}{c}{ResNet 101} \\ \cline{2-9}
                        & 1/32   & 1/16   & 1/8   & 1/4  & 1/32   & 1/16   & 1/8   & 1/4  \\ \hline
FPL+CPS w/o cutmix      & 0.95    & 0.9    & 0.9   & 0.9  & 0.95    & 0.95    & 0.95   & 0.9  \\
FPL+CPS w/ cutmix       & 0.9    & 0.85    & 0.85   & 0.85  & 0.9    & 0.85    & 0.85   & 0.85  \\
FPL+AEL                 & 0.95    & 0.95    & 0.9   & 0.9  & 0.9    & 0.9    & 0.85   & 0.85  \\ \hline
\end{tabular}\label{T city}
\vspace{-1.5em}
\end{table}

\begin{table}[htbp]\footnotesize
\setlength{\tabcolsep}{2.0mm}
  \centering
  \caption{\textbf{The setting of $T$ on VOC2012.}}
\begin{tabular}{l|ccc|ccc}
\hline
\multirow{2}{*}{Method} & \multicolumn{3}{c|}{ResNet 50} & \multicolumn{3}{c}{ResNet 101} \\ \cline{2-7}
                        & 1/16   & 1/8   & 1/4  & 1/16   & 1/8   & 1/4  \\ \hline
FPL+CPS w/o cutmix      & 0.9    & 0.9    & 0.9  & 0.95    & 0.9    & 0.9  \\
FPL+CPS w/ cutmix       & 0.95    & 0.9    & 0.9  & 0.9    & 0.9    & 0.9  \\
FPL+AEL                 & 0.95    & 0.95    & 0.95  & 0.95    & 0.9    & 0.9  \\ \hline
\end{tabular}\label{T voc}
\vspace{-1.5em}
\end{table}

\begin{table}[htbp]\footnotesize
\setlength{\tabcolsep}{3.0mm}
  \centering
  \caption{\textbf{The setting of $T$ on VOC2012 LowData.}}
\begin{tabular}{l|cccc}
\hline
Method            & 1/32 & 1/16 & 1/8 & 1/4 \\ \hline
FPL+CPS w/ cutmix & 0.95  & 0.85  & 0.85 & 0.85 \\ \hline
\end{tabular}\label{T voc lowdata}
\vspace{-1.5em}
\end{table}

\section{More Analysis}\label{sec:ana}
\subsection{Difference between FPL and negative learning}\label{sec:neg}
For uncertain unlabeled pixels, negative learning-based methods find their models always predict certainly that these pixels do not belong to some categories.
Hence, they treat the uncertain pixels as negative samples to those unlikely categories.
A commonly used paradigm sets a threshold (e.g., 0.2), and considers the classes for which the predicted probabilities are less than the threshold as negative categories~\cite{rizve2020defense}.
For clarity, we take the negative learning loss based on cross-entropy loss as the comparison object, since our method is also an extension of cross-entropy loss.
To unify the form, we denote the categories that do not belong to the negative categories as $\mathbb{Y}_{us}$.
Formulately, the negative loss $\mathcal{L}^{n}$ is:
\begin{equation}
\begin{aligned}
  \mathcal{L}^{n}(x_{us}) = \mathcal{L}_{us}^{n} = -\sum_{j \notin \mathbb{Y}_{us}}{\log(1-p_{us}^{j})}.
\end{aligned} \label{eq:negative loss}
\end{equation}
We see that this loss function requires the probabilities for negative categories to be small.
To further show the difference between $\mathcal{L}^{n}$ and our $\mathcal{L}^{f}$, we convert the $\mathcal{L}^{n}$ as:
\begin{equation}
\begin{aligned}
  &\mathcal{L}_{us}^{n} = -\sum_{j \notin \mathbb{Y}_{us}}{\log(1-p_{us}^{j})} = \sum_{j \notin \mathbb{Y}_{us}}{\log(\frac{1}{1-p_{us}^{j}})} \\
  & = \sum_{j \notin \mathbb{Y}_{us}}{\log(1+\frac{p_{us}^{j}}{1-p_{us}^{j}})} = \sum_{j \notin \mathbb{Y}_{us}}{\log(1+\frac{e^{z_{us}^{j}}}{\sum_{i \neq j}{e^{z_{us}^{i}}}}}) \\
  &\approx \sum_{j \notin \mathbb{Y}_{us}} ReLU(z_{us}^{j}-\underset{i \neq j} \max (z_{us}^{i})).
\end{aligned} \label{eq:negative loss}
\end{equation}
Eq.~\ref{eq:negative loss} shows that the negative loss implicitly increases the prediction for the top-1 pseudo label $\underset{i \neq j} \max (z_{us}^{i})$, indicating that it still corrupts the training of the model when pseudo labels are wrong.
Differently, our FPL desires to increase the predictions for all fuzzy positive categories in $\{z_{us}^{i}, i \in \mathbb{Y}_{us}\}$, hence we encourage their minimum $\min (z_{us}^{i})$ to learn the semantics of possible GT in them:
\begin{equation}
  \begin{aligned}
  \mathcal{L}^{f}_{us} \approx ReLU( \underset{j \notin \mathbb{Y}_{us}} \max(z_{us}^{j}) - \underset{i \in \mathbb{Y}_{us}} \min(z_{us}^{i}) ).
\end{aligned} \label{eq:loss function}
\end{equation}

Furthermore, we empirically demonstrate the superiority of FPL over the negative learning-based method.
Besides, we also evaluate the performance using a soft loss $\mathcal{L}^{s}$ with the soft label since it has similarities to FPL in softening pseudo labels, which is computed as:
\begin{equation}
\begin{aligned}
  \mathcal{L}^{s}(x_{us}) = \mathcal{L}_{us}^{s} = \sum_{i}^{C} q_{us}^{i}\log\frac{q_{us}^{i}}{p_{us}^{i}},
\end{aligned} \label{eq:soft loss}
\end{equation}
where $p_{us}$ is the predicted probability, and $q_{us}$ is the learning target.
Segmentation performances are shown in Table~\ref{tab:negative learning}, where `Nega.' represents the results obtained by negative loss $\mathcal{L}^{n}$, and `Soft.' represents the results obtained by soft loss $\mathcal{L}^{s}$.
In addition, U2PL~\cite{wang2022semi} introduces the idea of negative learning in the manner of contrastive learning, hence we also provide its performance here.
From Table~\ref{tab:negative learning}, we see our FPL model achieves the best performance, reflecting the superiority of FPL over other alternatives.
\begin{table}[htbp]\footnotesize
  \setlength{\tabcolsep}{4.0mm}
  \centering
  \caption{These results are obtained on Cityscapes using ResNet 101 as the backbone.}
\begin{tabular}{l|ccc}
\hline
{Method}
                        & 1/16      & 1/8      & 1/4      \\ \hline
CPS w/ cutmix                   & 74.72    & 77.62    & 78.93    \\
Soft.+ CPS w/ cutmix             & 73.19    & 77.43    & 78.75    \\
Nega.+ CPS w/ cutmix                   & 75.34    & 77.15    & 78.31    \\
U2PL~\cite{wang2022semi}                    & 74.90    & 76.48    & 78.51    \\ \hline
FPL+CPS w/ cutmix                     & \textbf{75.74}    & \textbf{78.47}    & \textbf{79.19} \\\hline
\end{tabular}\label{tab:negative learning}
\end{table}

\subsection{Gradient vanishing in K value selection strategy}\label{sec: gradient vanishing}
\begin{figure}[htbp]
\captionsetup{labelfont=bf}
\centerline{\includegraphics[scale=0.65]{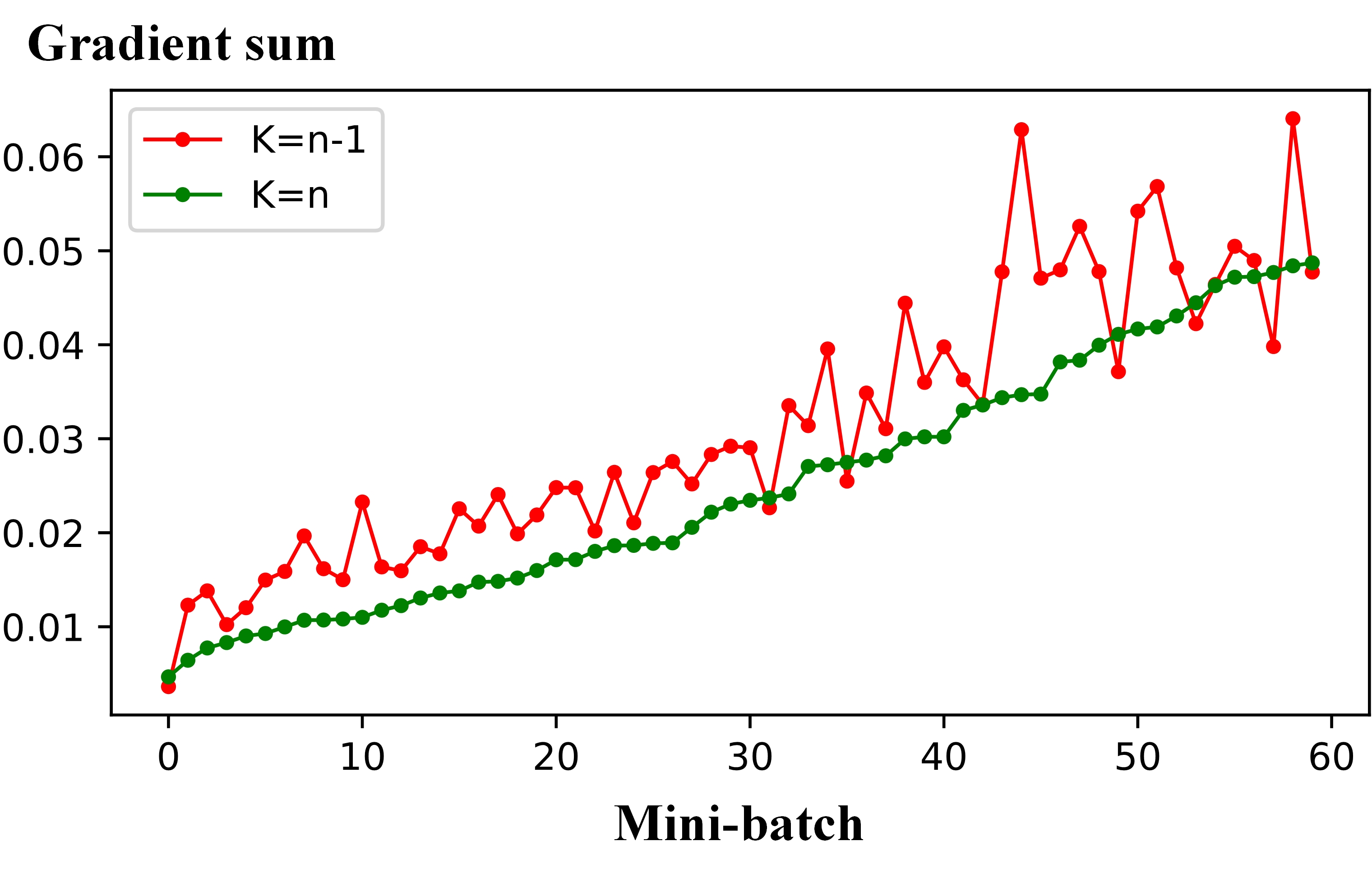}}
\caption{The gradient sum is $\sum_{j \notin \mathbb{Y}_{us}} \frac{\partial \mathcal{L}_{us}^{f}}{\partial z_{us}^{j}}$, and we sort these mini-batches by their gradient sum using $K=n$.
Here we only present examples with small (i.e., prone vanishing) gradients.
}
\label{fig:gradient_vanish}
\end{figure}

In our K value selection strategy, we select $K=n-1$ instead of $K=n$.
This practice is to alleviate the problem of gradient vanishing.
To explain this, we first perform an analysis in a simplified case where no perturbations are added in training, that is, the prediction that generates pseudo labels has the same distribution as the training prediction.
We further illustrate the actual gradient in training in Fig.~\ref{fig:gradient_vanish}.

\textbf{Analysis in simplified case.}
To discuss training gradient, we convert the gradients of $\mathcal{L}_{us}^{f}$ to probabilistic form:
\begin{equation}
\begin{aligned}
  &\frac{\partial \mathcal{L}_{us}^{f}}{\partial z_{us}^{i}} = \frac{-\sum_{j \notin \mathbb{Y}_{us}}p_{us}^{j}}{1+\sum_{j \notin \mathbb{Y}_{us}}p_{us}^{j} \times \sum_{i \in \mathbb{Y}_{us}}{\frac{1}{p_{us}^{i}}}} \times \frac{1}{p_{us}^{i}}\\
  &\frac{\partial \mathcal{L}_{us}^{f}}{\partial z_{us}^{j}} = \frac{\sum_{i \in \mathbb{Y}_{us}}\frac{1}{p_{us}^{i}}}{1+\sum_{j \notin \mathbb{Y}_{us}}p_{us}^{j} \times \sum_{i \in \mathbb{Y}_{us}}{\frac{1}{p_{us}^{i}}}} \times p_{us}^{j}.
\end{aligned} \label{eq:topk_gradient_in_p_pos}
\end{equation}
Here we only need to analyze the gradients of positive categories, because the absolute value of the gradient sum on the positive and negative categories are equal:
\begin{equation}
  \begin{aligned}
    &|\sum_{i\in \mathbb{Y}_{us}} \frac{\partial \mathcal{L}_{us}^{f}}{\partial z_{us}^{i}}| = \sum_{j \notin \mathbb{Y}_{us}} \frac{\partial \mathcal{L}_{us}^{f}}{\partial z_{us}^{j}}\\
    &= \frac{\sum_{j \notin \mathbb{Y}_{us}}{e^{z_{us}^{j}}} \times \sum_{i \in \mathbb{Y}_{us}}e^{-z_{us}^{i}}}{1+\sum_{j \notin \mathbb{Y}_{us}}{e^{z_{us}^{j}}} \times \sum_{i \in \mathbb{Y}_{us}}{e^{-z_{us}^{i}}}}.
  \end{aligned} \label{eq:pos_neg_gradient}
\end{equation}
From Eq.~\ref{eq:topk_gradient_in_p_pos}, we see that the $\frac{\partial \mathcal{L}_{us}^{f}}{\partial z_{us}^{i}}$ is close to 0 when its numerator (i.e. $\sum_{j \notin \mathbb{Y}_{us}}p_{us}^{j}$) is close to 0.
According to our K value selection strategy, the lower bound of $\sum_{j \notin \mathbb{Y}_{us}}p_{us}^{j}$ can be easily obtained.
If we choose $K_{us}=n$, then we get:
\begin{equation}
  \begin{aligned}
    \inf(\sum_{j \notin \mathbb{Y}_{us}}p_{us}^{j})=0,
  \end{aligned} \label{eq:kn_inf}
\end{equation}
where $\inf$ means the lower bound.
Eq.~\ref{eq:kn_inf} shows that it is possible for $\sum_{j \notin \mathbb{Y}_{us}}p_{us}^{j}$ to approach 0 causing the problem of gradient vanishing.
When setting $K_{us}$ an integer less than $n$ (i.e., $K_{us}=\left[\alpha \cdot n \right], 0<\alpha<1)$), we derive that:
\begin{equation}
  \begin{aligned}
    \inf(\sum_{j \notin \mathbb{Y}_{us}}p_{us}^{j})=\frac{\left[\alpha \cdot n \right]}{n-1}(1-T).
  \end{aligned} \label{eq:kn1_inf}
\end{equation}
Eq.~\ref{eq:kn1_inf} provides a lower bound for the numerator of $\frac{\partial \mathcal{L}_{us}^{f}}{\partial z_{us}^{i}}$, which alleviates the problem of gradient vanishing.
In practice, we use $K_{us}=n-1$ for all our experiments.

\textbf{Actual gradients in training.}
In actual training, the above inference will be deviated due to the influence of disturbance (e.g., data augmentation), but the conclusion still holds.
Considering that our model is also subject to a supervised loss $\mathcal{L}^{sup}$ except for the fuzzy positive loss $\mathcal{L}^{f}$.
A too-small gradient from $\mathcal{L}^{f}$ will lead the information of unlabeled data to be overwhelmed by the supervised loss.
We illustrate the actual gradients selecting $K=n-1$ and $K=n$ in Fig.~\ref{fig:gradient_vanish}.
It can be seen that $K=n$ brings a small training gradient while $K=n-1$ obtains a larger gradient in most mini-batches.

\subsection{More details of positive gradient score}\label{sec:score}
\begin{figure*}[htbp]
\captionsetup{labelfont=bf}
\centerline{\includegraphics[scale=0.53]{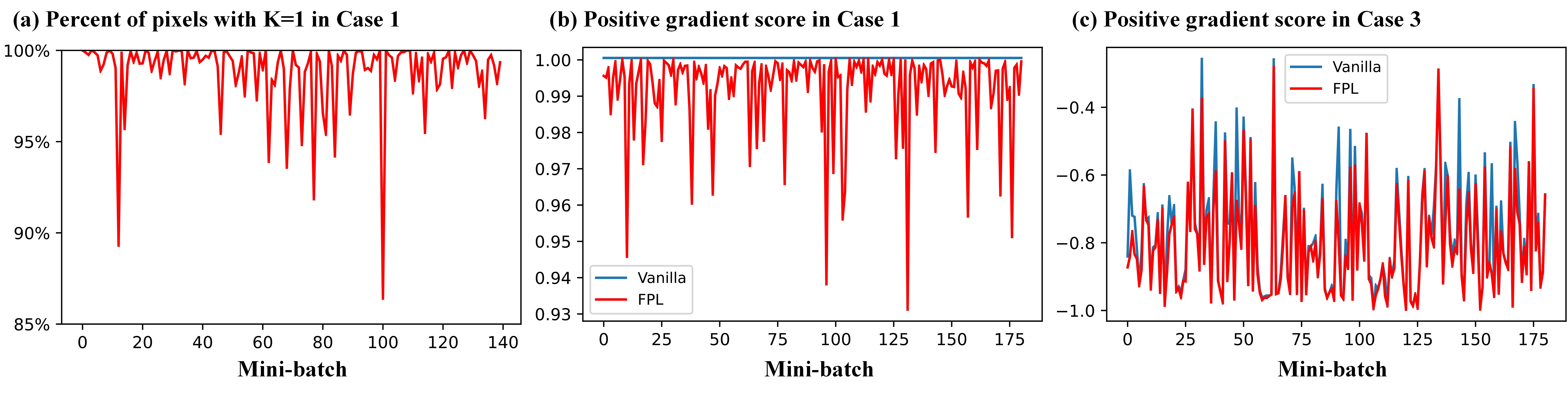}}
\caption{(a) The proportion of $K=1$ pixels in Case 1. (b) The positive gradient score of Case 1. (c) The positive gradient score of Case 3. This figure is plotted on VOC2012 with 1/8 labeled data.
}
\label{fig:grad_score_supp}
\vspace{-1.0em}
\end{figure*}

As shown in Fig.~\ref{fig:grad_score_supp} (a) and (b), we see that in Case 1, most pixels ($>$85\%) have $K=1$ and positive gradient score $R^{f}$ is very close to 1.
Besides, we see that $R^{f}$ is slightly lower than $R^{v}$ in Case 3.

\subsection{Gradient similarity between $G^{f}$ and $G^{ide}$}\label{sec:cos_sim}
In Case 2 of Sec.~3.4, though $\mathcal{L}^{f}$ encourages the GT prediction to increase which is better than existing $\mathcal{L}^{v}$, it also encourages the predictions for other positive categories to become larger.
Ideally, the cross-entropy loss using the unavailable ground truth only increases the GT prediction and suppresses the predictions for all other categories.
We name the gradient computed in this ideal situation as the ideal gradient $\bm{G^{ide}}$.

\begin{figure}[htbp]
\captionsetup{labelfont=bf}
\centerline{\includegraphics[scale=0.26]{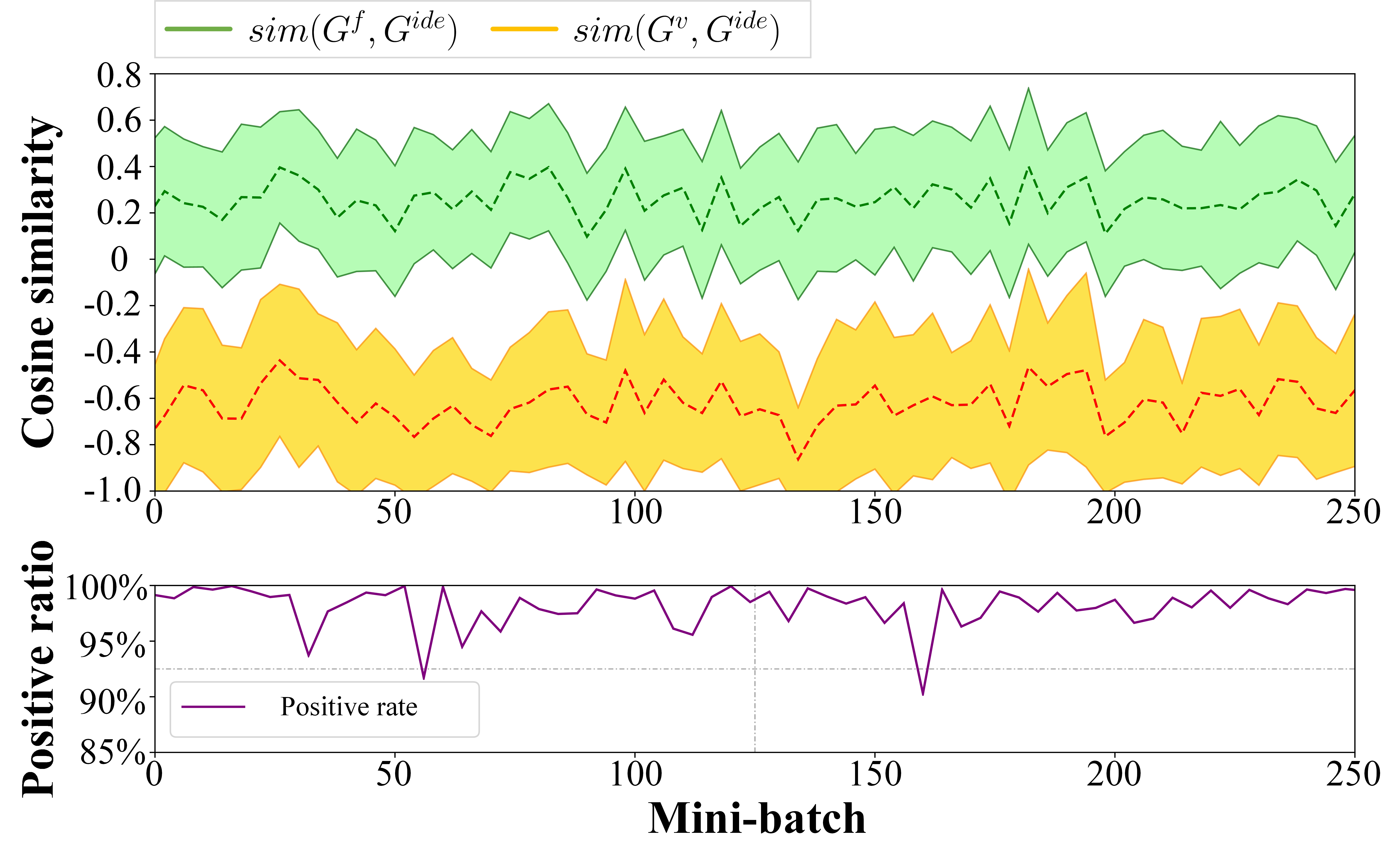}}
\caption{
The cosine similarity between ideal gradient $\bm{G^{ide}}$, fuzzy gradient $\bm{G^{f}}$, and vanilla gradient $\bm{G^{v}}$.
Also, we present the positive rate of the similarity between $\bm{G^{ide}}$ and $\bm{G^{f}}$ computed on pixels in each minibatch.
This figure is counted on VOC2012 with 1/8 labeled data using the CPS~\cite{chen2021semi} framework.
}
\label{fig:analysis_cos_sim}
\vspace{-1.0em}
\end{figure}

Here, we propose to use the cosine similarity between the ideal gradient vector $\bm{G^{ide}}$ and our fuzzy gradient vector $\bm{G^{f}}$ brought by $\mathcal{L}^{f}$ to further analyze our FPL in Case 2.
If the cosine similarity is greater than 0, it means the projection of $\bm{G^{f}}$ on $\bm{G^{ide}}$ is positive, indicating $\bm{G^{f}}$ makes our model go further in the ideal direction.
For comparison, we also present the cosine similarity between the gradient vector of the vanilla method $\bm{G^{v}}$ and the ideal gradient $\bm{G^{ide}}$.
Due to the complexity of predicted probability, the relationship between the cosine similarity $sim (\bm{G^{f}}, \bm{G^{ide}}) = \frac{\bm{G^{f}} \cdot \bm{G^{ide}}}{|\bm{G^{f}}| |\bm{G^{ide}}|}$ and 0 is not mathematically absolute.
Therefore, we count $sim (\bm{G^{f}}, \bm{G^{ide}})$ and $sim (\bm{G^{v}}, \bm{G^{ide}})$ quantitatively.
As shown in Fig.~\ref{fig:analysis_cos_sim}, we first observe that the positive rates of $sim (\bm{G^{f}}, \bm{G^{ide}})$ are more than 90\% in all mini-batches, which indicates that $\bm{G^{f}}$ makes our model go further in the ideal direction in most cases.
Second, we see that the $sim (\bm{G^{f}}, \bm{G^{ide}})$ is greater than the $sim (\bm{G^{v}}, \bm{G^{ide}})$, which means our fuzzy gradient $\bm{G^{f}}$ is closer to the ideal gradient $\bm{G^{ide}}$ than the gradient from vanilla method $\bm{G^{v}}$.

\begin{figure}[htbp]
\centering
\includegraphics[scale=0.34]{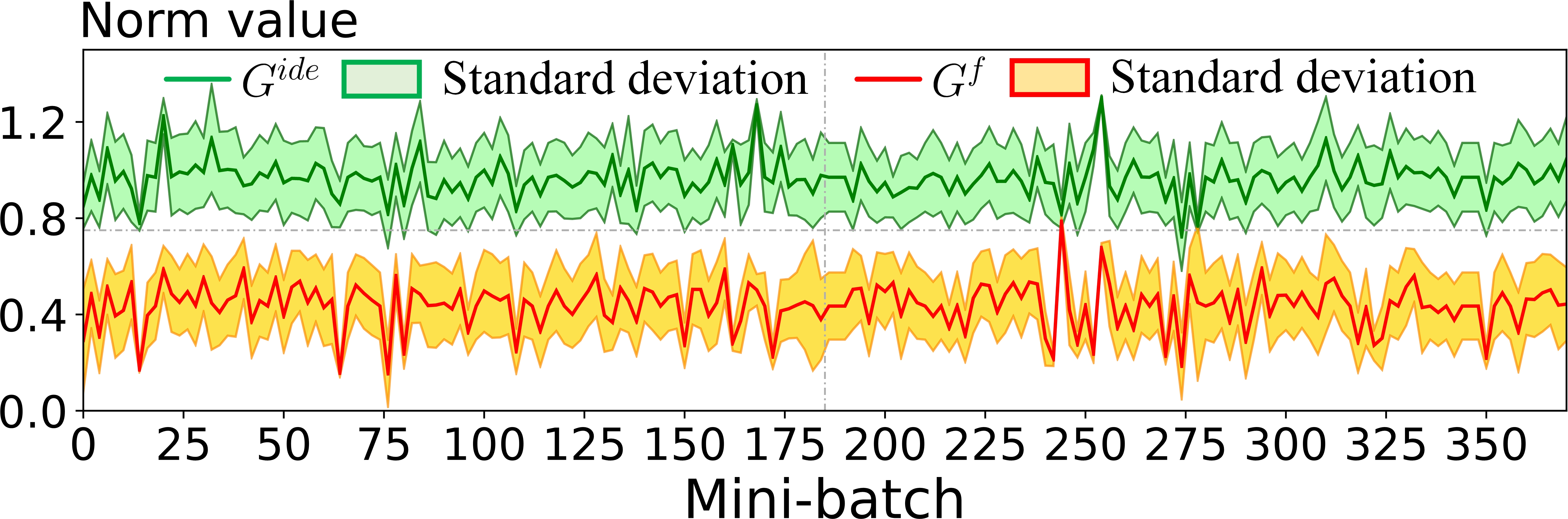}
\caption{
The norms of $\bm{G^{ide}}$ and $\bm{G^{f}}$ counted on VOC2012 with 1/8 labeled data using the CPS~\cite{chen2021semi} framework.
}\label{fig:norms}
\vspace{-1.0em}
\end{figure}

\textbf{The norms of $\bm{G^{f}}$ and $\bm{G^{ide}}$.}
The $sim (\bm{G^{f}}, \bm{G^{ide}})$ only reflects that the angle between our fuzzy gradient and the ideal gradient is a mostly acute angle.
But the norms of $\bm{G^{f}}$ and $\bm{G^{ide}}$ also affects optimization of our model.
If the norm of $\bm{G^{f}}$ is much larger than that of $\bm{G^{ide}}$, it will cause $\bm{G^{f}}$ over-optimize our model, hence even if their angle is small, it will also be detrimental to optimization.
We prove that the norms of $\bm{G^{f}}$ and $\bm{G^{ide}}$ are both range of $[0,\sqrt{2}]$:
\begin{equation}
  \begin{aligned}
    &|N^{ide}| = \sqrt{(g^{1})^{2} + (g^{2})^{2} + ... + (g^{C})^{2}} \\
    &= \sqrt{(g^{y})^{2}+\sum_{i \neq y}(g^{i})^{2}}= \sqrt{(\sum_{i \neq y}g^{i})^{2} + \sum_{i \neq y}(g^{i})^{2}}\\
    &\leq\sqrt{ 2 (\sum_{i \neq y}g^{i})^{2}} \leq \sqrt{2}
  \end{aligned} \label{eq:norms_ide}
\end{equation}
\begin{equation}
  \begin{aligned}
    &|N^{f}| = \sqrt{(g^{1})^{2} + (g^{2})^{2} + ... + (g^{C})^{2}}\\
    &= \sqrt{\sum_{i \in \mathbb{Y}} (g^{i})^{2}+\sum_{i \notin \mathbb{Y}}(g^{j})^{2}} \leq \sqrt{(\sum_{i \in \mathbb{Y}} g^{i})^{2} + (\sum_{i \notin \mathbb{Y}}g^{j})^{2}}\\
    &= \sqrt{2 \times (\sum_{i \notin \mathbb{Y}}g^{j})^{2}} \leq \sqrt{2}.
  \end{aligned} \label{eq:norms_p}
\end{equation}
Quantitatively, we provide the norms of $\bm{G^{f}}$ and $\bm{G^{ide}}$ in Fig.~\ref{fig:norms}.
We see that the two norms are close and the norm of $\bm{G^{f}}$ is smaller than that of $\bm{G^{ide}}$, which means that our FPL won't bring the problem of over-optimization.

\section{More Ablation Studies}\label{sec:abla}
\subsection{Adaptive Weight}\label{sec: adaptive weight}
\begin{figure}[htbp]
\centerline{\includegraphics[scale=0.5]{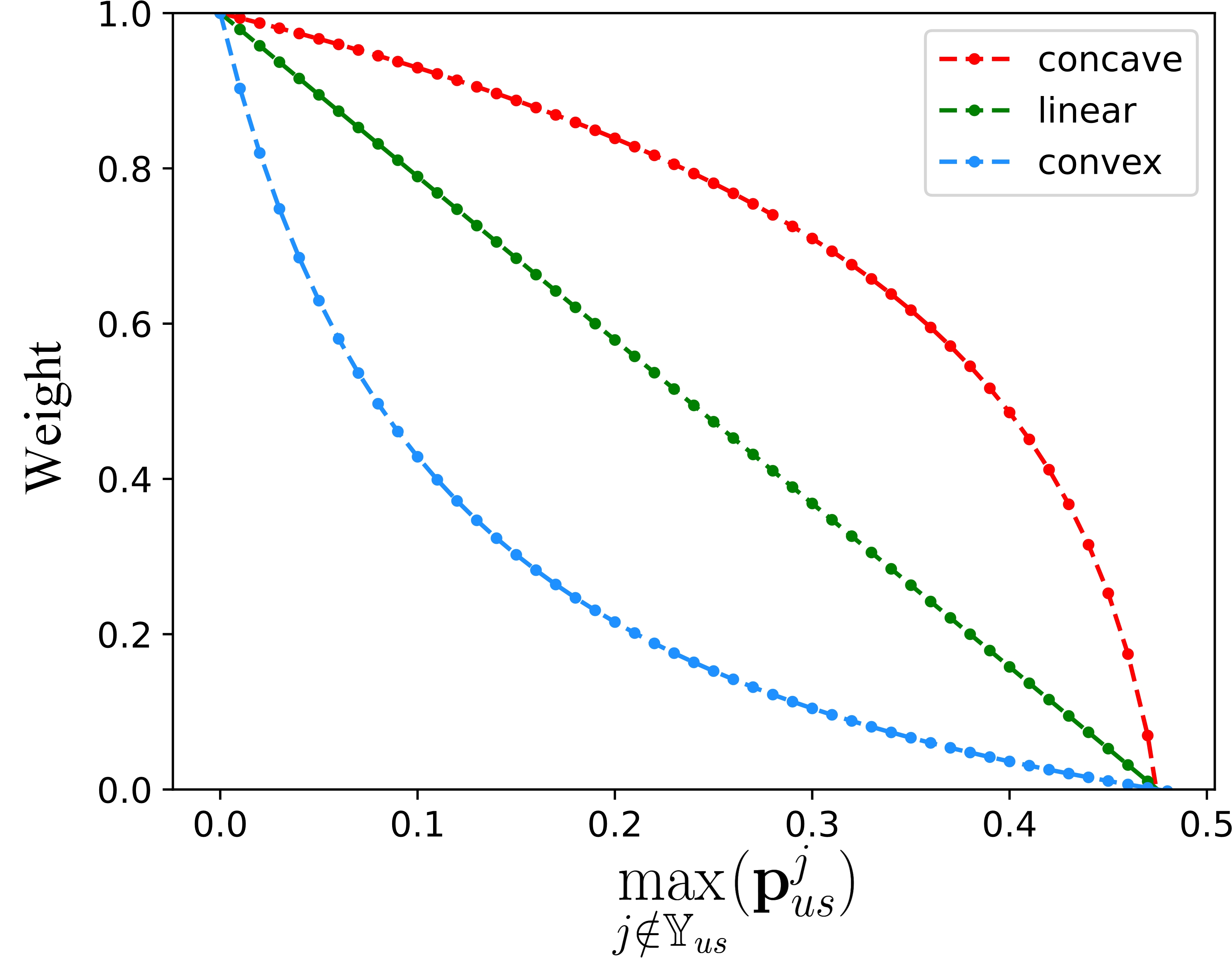}}
\caption{
Examples of convex, linear, and concave decreasing functions.
}
\label{fig:adaptive_weight_function}
\end{figure}

\begin{figure}[htbp]
\centerline{\includegraphics[scale=0.46]{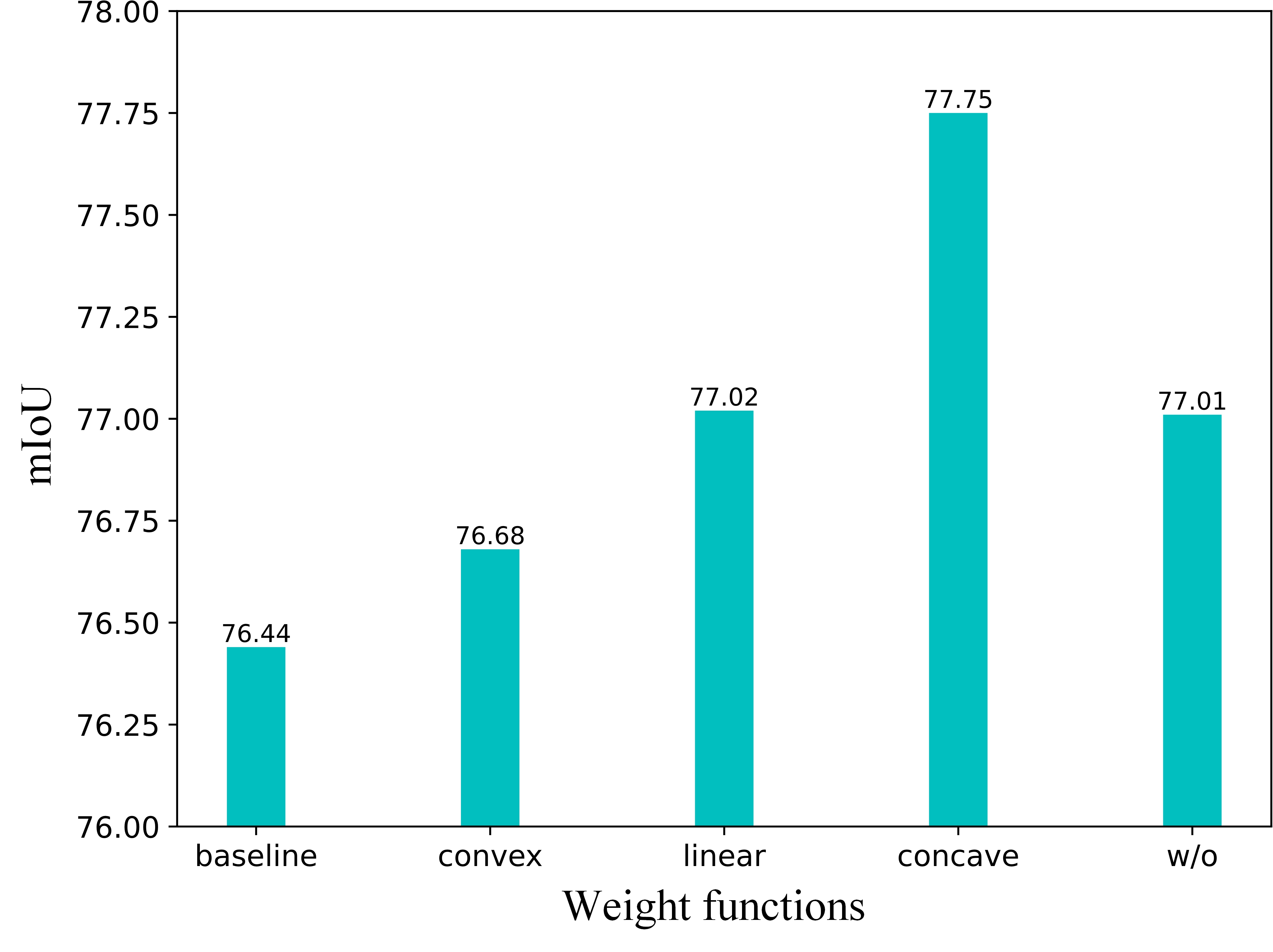}}
\caption{
Performance of our FPL+CPS w/ CutMix model using different adaptive weights.
The `baseline' and `w/o' represent the baseline CPS model and our FPL model trained without adaptive weight, respectively.
These results are obtained on VOC2012 with 1/8 labeled data.
}
\label{fig:adaptive_weight_performance}
\end{figure}

In Sec.~3.3 of our manuscript, we show that the adaptive weight function should be inversely proportional to $\underset{j \notin \mathbb{Y}_{us}} \max(p_{us}^{j})$.
Here we provide an experiment showing that the used concave decreasing function performs better than linear or convex decreasing functions.
The function curves are illustrated in Figure~\ref{fig:adaptive_weight_function}, which are plotted in the setting of $T=0.95$ and $K=2$.
And the formulas of convex and linear functions are expressed as:
\begin{equation}
  \begin{aligned}
    &w_{convex} = \frac{\frac{\sum_{i \in \mathbb{Y}_{us}}{p_{us}^{i}}}{K_{us}} - \underset{j \notin \mathbb{Y}_{us}} \max(p_{us}^{j})}{\frac{\sum_{i \in \mathbb{Y}_{us}}{p_{us}^{i}}}{K_{us}}+4*\underset{j \notin \mathbb{Y}_{us}} \max(p_{us}^{j})}\\
    &w_{linear} = \frac{\frac{\sum_{i \in \mathbb{Y}_{us}}{p_{us}^{i}}}{K_{us}} - \underset{j \notin \mathbb{Y}_{us}} \max(p_{us}^{j})}{\frac{\sum_{i \in \mathbb{Y}_{us}}{p_{us}^{i}}}{K_{us}}}.
  \end{aligned} \label{eq:other weight function}
\end{equation}
The segmentation performances are shown in Fig.~\ref{fig:adaptive_weight_performance}, where we also provide the comparison of the model trained without adaptive weight, that is, the weights for all pixels are the same as 1.
We see that the used convex function performs better than other alternatives.

\begin{table}[thbp]\footnotesize
\setlength{\tabcolsep}{1.8mm}
\centering
\caption{\textbf{Ablation study on K value selection strategy.} Results are obtained on VOC2012 and Cityscapes with 1/16 labeled data.\label{tab:K value selection strategy}}
\begin{tabular}{c|ccc|cc}
\hline
K strategy                            & K=3 & K=2 & Step & Ours(K=n) & Ours(K=n-1) \\ \hline
FPL+CPS & 56.71     & 61.09     & 66.39      & 65.98     & \textbf{68.67}   \\ \hline
\end{tabular}
\end{table}

\subsection{K value selection strategy}\label{sec: K strategy}
Here we evaluate the superiority of the proposed K value selection strategy by comparing our strategy with a fixed K value strategy and a step-decay K value strategy.
The step decay strategy is to initialize the K value to 3 and decrease K by one every 1/3 of the total training epochs.
In addition, we also verify that $K=n-1$ is better than $K=n$ in our K value selection strategy.
The results are shown in Table~\ref{tab:K value selection strategy}.
We see that fixed K value results in a large degradation in the performance of FPL since a fixed K value causes the model to produce high-entropy predictions, making it difficult to obtain accurate classifications.
For the step decay K value strategy, it achieves better results than fixed K values, because it could reduce the K value during training to obtain low-entropy classifications.
However, it is still worse than our proposed strategy since it makes K values the same for all pixels, ignoring their difference in the learning progress.
In contrast, our method adaptively chooses the K value for each pixel according to its predicted probability distribution.
We also see that $K=n-1$ is better than $K=n$ in our K value selection strategy.
This is because selecting $K=n-1$ alleviates the gradient vanishing problem.

\begin{figure*}[htbp]
\centerline{\includegraphics[scale=0.38]{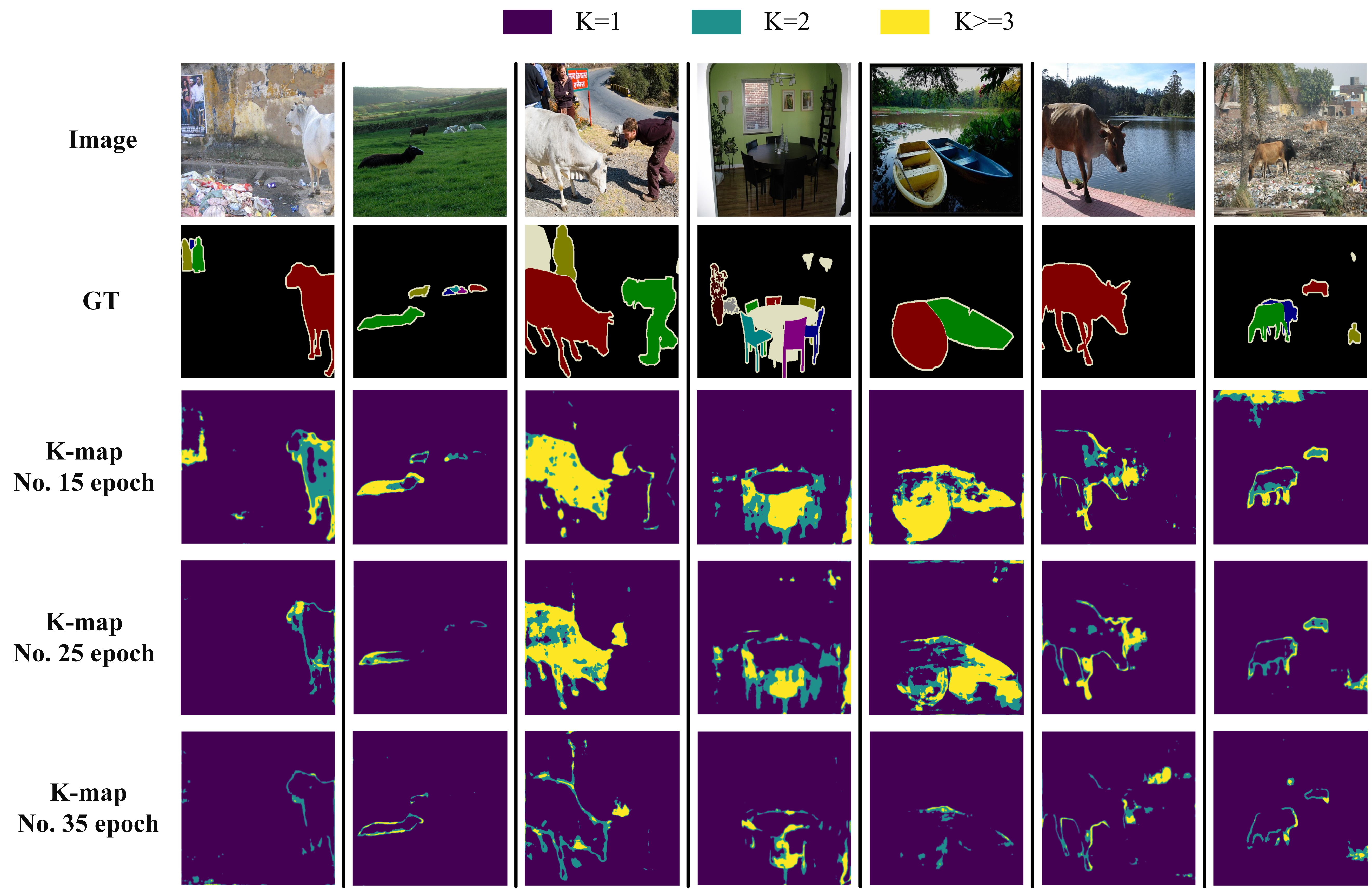}}
\caption{
The visualization of K values.
This figure is plotted on the VOC2012 with 1/16 labeled data using FPL+CPS w/ CutMix as the training method.
}
\label{fig:k map half}
\end{figure*}

\begin{figure*}[htbp]
\centerline{\includegraphics[scale=0.38]{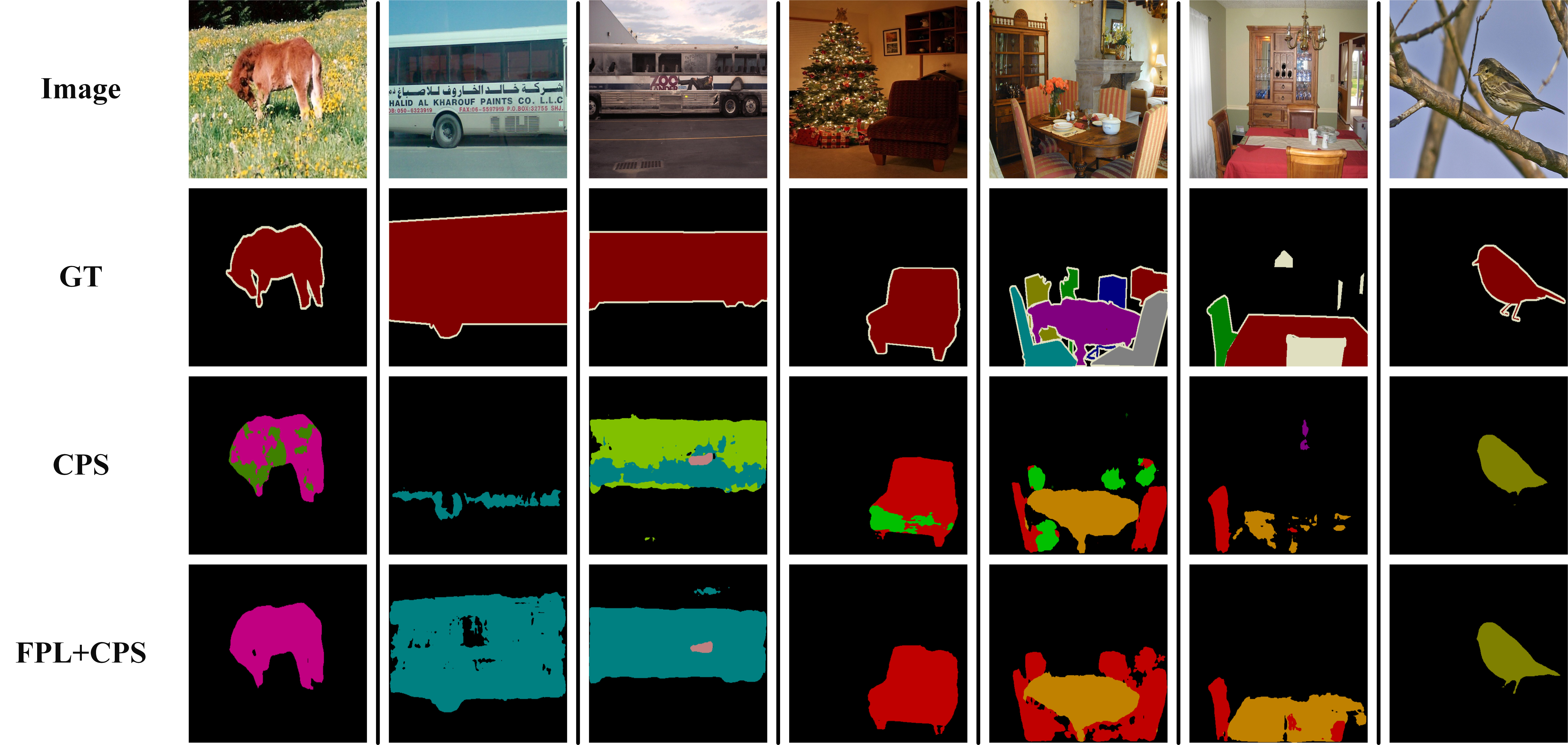}}
\caption{
These segmentation results are obtained on the VOC2012 dataset with 1/16 labeled data.
For clarity, we show ground truth (GT) in the form of instance labels, and the predictions of CPS and FPL+CPS are presented in the form of semantic labels.
}
\label{fig:result_map}
\end{figure*}

\section{Limitations}\label{sec:limitation}
Though works well, FPL has the limitation of high time complexity since it requires assigning a K value to each pixel.
From Eq.~(7) of our manuscript, we see that the time complexity of computing $\mathcal{L}^{f}$ is $O(C)$ when the $\mathbb{Y}$ is determined, where $C$ is the number of classes.
For vanilla $\mathcal{L}^{v}$, it is a special case of $\mathcal{L}^{f}$ when the $K$ is fixed to 1, hence the time complexity of original $\mathcal{L}^{v}$ for one pixel is $O(C)$.
When it comes to $\mathcal{L}^{f}$, we additionally need to decide the K value for each pixel of which the time complexity is $O(K)$ since it needs K times additions and K times comparisons.
Hence, the time complexity of computing $\mathcal{L}^{f}$ is $O(KC)$ which is K times of computing the original $\mathcal{L}^{v}$.
We also quantitatively provide the seconds of training our FPL in practice.
As shown in Table~\ref{tab:time limitation}, FPL brings about 15\% additional training cost.

\begin{table}[htbp]\footnotesize
  \setlength{\tabcolsep}{5mm}
  \caption{\textbf{Seconds per epoch.} These statistics are measured using 8 Tesla V100 GPUs under the setting of 1/8 labeled data with ResNet 101 baseline.}
  \centering
\begin{tabular}{l|cc}
\hline
Method  & Cityscapes & VOC2012  \\ \hline
AEL~\cite{hu2021semi}     & 730s              & 835s         \\
FPL+AEL & 820s             & 985s           \\ \hline
\end{tabular}\label{tab:time limitation}
\end{table}

\section{Visualization}\label{sec:vis}
We present more samples of K value maps during training in Fig.~\ref{fig:k map half}.
And we illustrate some examples of our segmentation results in Fig.~\ref{fig:result_map}.

\end{document}